\documentclass[11pt]{article} 
\usepackage{etoolbox}

 \usepackage[letterpaper, left=1in, right=1in, top=1in, bottom=1in]{geometry}

\usepackage[colorlinks=true, linkcolor=blue!70!black, citecolor=blue!70!black]{hyperref}
\usepackage{microtype}

\usepackage{natbib}
\bibpunct{(}{)}{;}{a}{,}{,}

\usepackage{amsthm}
\usepackage{mathtools}
\usepackage{amsmath}
\usepackage{bbm}
\usepackage{amsfonts}
\usepackage{amssymb}

\usepackage{xpatch}
\usepackage{pifont}
\usepackage{array}
\usepackage{booktabs}
\usepackage{floatrow}
\newfloatcommand{capbtabbox}{table}[][\FBwidth]
\usepackage{blindtext}
\usepackage{caption}

\theoremstyle{definition}  
\theoremstyle{theorem}
\newtheorem{lemma}{Lemma}
\newtheorem{theorem}{Theorem}

\newtheorem{assumption}{Assumption}
\newtheorem{corollary}{Corollary}

\newtheorem{definition}{Definition}

\xpatchcmd{\proof}{\itshape}{\normalfont\proofnameformat}{}{}
\newcommand{\proofnameformat}{\bfseries}

\usepackage{prettyref}
\newcommand{\pref}[1]{\prettyref{#1}}

\newcommand{\savehyperref}[2]{\texorpdfstring{\hyperref[#1]{#2}}{#2}}
\newrefformat{eq}{\savehyperref{#1}{\textup{(\ref*{#1})}}}
\newrefformat{ineq}{\savehyperref{#1}{\textup{(\ref*{#1})}}}
\newrefformat{eqn}{\savehyperref{#1}{Equation~\ref*{#1}}}
\newrefformat{obs}{\savehyperref{#1}{Observation~\ref*{#1}}}
\newrefformat{lem}{\savehyperref{#1}{Lemma~\ref*{#1}}}
\newrefformat{def}{\savehyperref{#1}{Definition~\ref*{#1}}}
\newrefformat{thm}{\savehyperref{#1}{Theorem~\ref*{#1}}}
\newrefformat{fact}{\savehyperref{#1}{Fact~\ref*{#1}}}
\newrefformat{tab}{\savehyperref{#1}{Table~\ref*{#1}}}
\newrefformat{table}{\savehyperref{#1}{Table~\ref*{#1}}}
\newrefformat{line}{\savehyperref{#1}{Line~\ref*{#1}}}
\newrefformat{corr}{\savehyperref{#1}{Corollary~\ref*{#1}}}
\newrefformat{cor}{\savehyperref{#1}{Corollary~\ref*{#1}}}
\newrefformat{sec}{\savehyperref{#1}{Section~\ref*{#1}}}
\newrefformat{app}{\savehyperref{#1}{Appendix~\ref*{#1}}}
\newrefformat{ass}{\savehyperref{#1}{Assumption~\ref*{#1}}}
\newrefformat{ex}{\savehyperref{#1}{Example~\ref*{#1}}}
\newrefformat{fig}{\savehyperref{#1}{Figure~\ref*{#1}}}
\newrefformat{alg}{\savehyperref{#1}{Algorithm~\ref*{#1}}}
\newrefformat{rem}{\savehyperref{#1}{Remark~\ref*{#1}}}
\newrefformat{conj}{\savehyperref{#1}{Conjecture~\ref*{#1}}}
\newrefformat{prop}{\savehyperref{#1}{Proposition~\ref*{#1}}}
\newrefformat{proto}{\savehyperref{#1}{Protocol~\ref*{#1}}}
\newrefformat{prob}{\savehyperref{#1}{Problem~\ref*{#1}}}
\newrefformat{claim}{\savehyperref{#1}{Claim~\ref*{#1}}}
\newrefformat{tbl}{\savehyperref{#1}{Table~\ref*{#1}}}

\usepackage{enumitem}      

\newcommand\numberthis{\addtocounter{equation}{1}\tag{\theequation}}
\allowdisplaybreaks

\DeclarePairedDelimiter{\abs}{\lvert}{\rvert} 
\DeclarePairedDelimiter{\brk}{[}{]}
\DeclarePairedDelimiter{\crl}{\{}{\}}
\DeclarePairedDelimiter{\prn}{(}{)}
\DeclarePairedDelimiter{\nrm}{\|}{\|}
\DeclarePairedDelimiter{\tri}{\langle}{\rangle}

\let\Pr\undefined

\DeclareMathOperator{\En}{\mathbb{E}}

\DeclareMathOperator{\Pr}{Pr}

\DeclareMathOperator*{\argmin}{argmin} 

\newcommand{\ldef}{\vcentcolon=}
\newcommand{\rdef}{=\vcentcolon}

\newcommand{\wt}[1]{\widetilde{#1}}
\newcommand{\wh}[1]{\widehat{#1}}

\def\ddefloop#1{\ifx\ddefloop#1\else\ddef{#1}\expandafter\ddefloop\fi}
\def\ddef#1{\expandafter\def\csname bb#1\endcsname{\ensuremath{\mathbb{#1}}}}
\ddefloop ABCDEFGHIJKLMNOPQRSTUVWXYZ\ddefloop
\def\ddefloop#1{\ifx\ddefloop#1\else\ddef{#1}\expandafter\ddefloop\fi}
\def\ddef#1{\expandafter\def\csname b#1\endcsname{\ensuremath{\mathbf{#1}}}}
\ddefloop ABCDEFGHIJKLMNOPQRSTUVWXYZ\ddefloop
\def\ddef#1{\expandafter\def\csname c#1\endcsname{\ensuremath{\mathcal{#1}}}}
\ddefloop ABCDEFGHIJKLMNOPQRSTUVWXYZ\ddefloop
\def\ddef#1{\expandafter\def\csname h#1\endcsname{\ensuremath{\widehat{#1}}}}
\ddefloop ABCDEFGHIJKLMNOPQRSTUVWXYZabcdefghijklmnopqrsuvwxyz\ddefloop    
\def\ddef#1{\expandafter\def\csname hc#1\endcsname{\ensuremath{\widehat{\mathcal{#1}}}}}
\ddefloop ABCDEFGHIJKLMNOPQRSTUVWXYZ\ddefloop
\def\ddef#1{\expandafter\def\csname t#1\endcsname{\ensuremath{\widetilde{#1}}}}
\ddefloop ABCDEFGHIJKLMNOPQRSTUVWXYZ\ddefloop
\def\ddef#1{\expandafter\def\csname tc#1\endcsname{\ensuremath{\widetilde{\mathcal{#1}}}}}
\ddefloop ABCDEFGHIJKLMNOPQRSTUVWXYZ\ddefloop

\usepackage{tikz}

\newcommand{\grad}{\nabla}

\newcommand{\proman}[1]{\prn*{\romannumeral #1}}

\newcommand{\overleq}[1]{\overset{ #1}{\leq{}}}

\newcommand{\newer}[1]{{#1}}

\newcommand{\ignore}[1]{}

\renewcommand{\epsilon}{\varepsilon}

\renewcommand{\bA}{\bar{A}}

\newcommand{\newest}[1]{#1}

\usepackage{xcolor}

\usepackage{color-edits}
\addauthor{as}{red} 
\addauthor{gk}{purple}  
\addauthor{jd}{orange}
\addauthor{th}{blue}   

\usepackage{mathrsfs}

\usepackage{longtable} 
\usepackage{import}
\usepackage{enumitem} 
\usepackage{algorithm}
\usepackage[noend]{algpseudocode} 
\renewcommand{\algorithmicrequire}{\textbf{Input:}}
\algblockdefx[class]{Class}{EndClass}[1]{\textbf{object} \textsc{#1}:}{\textbf{end object}}
\makeatletter 
\ifthenelse{\equal{\ALG@noend}{t}} 
  {\algtext*{EndClass}}
  {}
\makeatother
\usepackage{amsmath}
\usepackage{makecell}
\usepackage{enumitem}  
\usepackage{bbm} 
\usepackage{bbold}
\usepackage{xspace}
\usepackage[scaled=.9]{helvet}
\usepackage{booktabs} 
\usepackage[export]{adjustbox}
\title{Remember What You Want to Forget: \\ Algorithms for Machine Unlearning}

\newcommand\blfootnote[1]{%
  \begingroup
  \renewcommand\thefootnote{}\footnote{#1}%
  \addtocounter{footnote}{-1}%
  \endgroup
}

\author{Ayush Sekhari$^{\dag}$ $\quad$ Jayadev Acharya$^{\ddag}$ $\quad$ Gautam Kamath$^{*}$ $\quad$ Ananda Theertha Suresh$^{\S}$}

\date{}  

\begin{document}

\maketitle 
\begin{abstract} 
We study the problem of unlearning datapoints from a learnt model. The learner first receives a dataset $S$ drawn i.i.d.\ from an unknown distribution, and outputs a model $\wh w$ that performs well on  unseen samples from the same distribution. However, at some point in the  future, any training datapoint $z \in S$ can request to be unlearned, thus prompting the learner to modify its output model while still ensuring the same accuracy guarantees.  We initiate a rigorous study of generalization in machine unlearning, where the goal is to perform well on previously unseen datapoints. Our focus is on both computational and storage complexity. 

For the setting of convex losses, we provide an unlearning algorithm that can unlearn up to $O(n/d^{1/4})$ samples, where $d$ is the problem dimension. In comparison, in general, differentially private learning (which implies unlearning) only guarantees deletion of $O(n/d^{1/2})$ samples. This demonstrates a novel separation between differential privacy and machine unlearning. 
\end{abstract} 

\blfootnote{$^{\dagger}$ Cornell University, \texttt{as3663@cornell.edu}.} 
\blfootnote{$^{\ddag}$ Cornell University, \texttt{acharya@cornell.edu}.} 
\blfootnote{$^{*}$ University of Waterloo, \texttt{g@csail.mit.edu}.} 
\blfootnote{$^{\S}$ Google Research, NY \texttt{theertha@google.com}.} 

\section{Introduction}
\label{sec:intro}

\newest{Many organizations and companies employ user data to train machine
learning models for a wide array of applications, ranging from movie
recommendations to health care.} 
While some of these organizations 
allow users to withdraw their consent from \newest{their} data being used (at which
point the organization will delete the user's data), less savory
businesses might covertly retain user data.  Given the potential for
misuse, legislators worldwide have wisely introduced laws that mandate
user data deletion upon request. {These} include the European
Union's \citet{GDPR16} (GDPR), the California Consumer Privacy Act
(CCPA), and Canada's proposed Consumer Privacy Protection Act (CPPA). 

There is some natural ambiguity present in these guidelines.  Is
it sufficient to simply delete the user's data, or must one also take 
action on machine learning systems that used this data for training?
Indeed, by now, privacy researchers are well-aware that user data may
be extracted from trained machine learning models
(e.g.,~\citet{ShokriSSS17,CarliniLEKS19}).  In a potentially landmark
decision, the Federal Trade Commission recently ordered a company to
delete not only data from users who deleted their accounts, but also
models and algorithms derived from this data~\citep{FTC21}.  This
suggests that organizations have an obligation to retrain any machine
learning models \newer{after} excluding users whose data has been deleted.

However, na\"ively retraining models after every deletion request would
be prohibitively expensive: training modern machine learning models
may take weeks, and use resources of value in the millions.
One could instead imagine more careful methods, which attempt to
excise the required datapoints from the model: crucially,
\emph{without} incurring the cost of retraining from scratch.  This
notion is called \emph{machine unlearning}.  The goal would be to
obtain a model which is \emph{identical} \newer{to the alternative model that
would be obtained when trained on the dataset after removing the
points that need to be forgotten}.  This requirement is rather strong: \citet{GinartGVZ19} 
proposed a relaxed notion of deletion, in which the model must only be
\emph{close} to the alternative, where closeness is defined in a way
reminiscent of differential privacy~\citep{DworkMNS06, DworkKMMN06}
(our variant of this notion is described in
\pref{def:unlearning}).  This relaxation has inspired the
design of several efficient algorithms for data deletion from machine
learning models \citep{GuoGHV20,IzzoSCZ21,NeelRS21, ullah2021machine}. 

As mentioned before, one na\"ive strategy involves retraining the model from scratch, sans the deleted datapoints.
When the training dataset is large, this approach is undesirable for several reasons.
First, it is computationally very expensive. 
Even iteration over the training data can be too costly, let alone training a new model on it.
Second, preserving the entire training dataset consumes a significant amount of storage.\footnote{Orthogonal to storage constraints, an additional issue is that government regulations may restrict the learner from storing raw user data for extended periods of time due to privacy concerns. 
However, we focus on storage as it captures undesirability of a wider range of unsatisfactory solutions.}

Another straightforward approach involves model checkpointing, in which the learner preemptively stores backup models in which certain points have been excluded.
While this strategy makes it easy to quickly return an appropriate backup model upon receiving a deletion request, the downside is that one typically has to store a number of additional models which scales with the training data size, which may be prohibitively large.
As we can see from these examples, computational and storage complexity are two vital metrics when designing a machine unlearning algorithm.

Finally, while there has recently been a wealth of results in machine unlearning, all of it has focused on the core problem of \newest{\emph{empirical risk minimization}, where the goal is to minimize the training loss}.
  However, to fulfil the promise of machine learning, we desire algorithms that can \emph{generalize} to previously unseen test data. 
  Motivated simultaneously by all of these concerns, our goal is to address the following question:

\begin{center} 
\textit{How do we design resource-efficient machine unlearning algorithms which generalize?}  
\end{center} 
\paragraph{Our contributions.} We initiate a new line of inquiry in machine 
unlearning:  
\begin{enumerate}[label=$\bullet$, leftmargin=8mm] 
\item We investigate generalization properties of unlearning algorithms, in particular asking: how many samples can we unlearn while still ensuring good performance on unseen test data? 
In comparison, prior work focused on the empirical training loss only. 

\item We consider machine unlearning simultaneously under storage constraints as well as the previously studied computation constraints. 
  Unlike prior work, our algorithms do not require the training data to be available to the unlearning algorithm when deleting samples. 

\item A clean approach for unlearning {is} to ignore which particular samples are being unlearnt and directly apply 
known algorithms and guarantees from differential privacy (DP). We show a strict separation between DP and machine unlearning.

In particular, algorithms based on DP can delete at most $\widetilde{\Theta}(n/\sqrt{d})$ samples while still retaining test loss performance, where $d$ denotes the  dimension of the problem. On the other hand, we provide efficient unlearning algorithms that take into account the particular samples to be unlearnt and show that we can delete up to
$\widetilde{O}(n / d^{1/4})$ samples, thus giving a quadratic improvement
in terms of dependence of $d$ over DP.  Our results apply to both strongly 
convex and convex loss functions. 
\end{enumerate} 

\subsection{Related work} 
\label{sec:related}
\citet{CaoY15} introduced the term ``machine unlearning,'' and gave efficient deterministic algorithms for exact unlearning in certain settings.  This definition requires an algorithm to have identical
outputs on a dataset after deleting a point, and if that point was 
never inserted. However, their algorithms are restricted to very structured problems only.  \citet{BourtouleCCJTZLP21} provide unlearning
algorithms using a sharding-based strategy, though in a weaker
unlearning model (requiring only that it be possible that the output
may have arisen), and without error guarantees.

\citet{GinartGVZ19}
introduced the probabilistic notion of unlearning, inspired by
differential privacy \citep{DworkKMMN06, DworkMNS06}. Their definition requires the output distribution of the unlearning algorithm to be similar to the output distribution obtained by running the learning algorithm on the dataset without the deleted points. Several recent
works~\citep{GuoGHV20,IzzoSCZ21,NeelRS21, ullah2021machine} {provide} theoretical error
guarantees for various problem settings under this probabilistic notion of unlearning. While our unlearning setup is closely related to that of \citet{GinartGVZ19} and in the related works, there are two major differences.

First, the prior work focuses on empirical risk minimization \citep{GuoGHV20,IzzoSCZ21,NeelRS21, ullah2021machine}. In their setup, the goal of the unlearning algorithm is to find approximate minimizers of the empirical loss on the remaining training dataset after deleting samples. In comparison, our focus in this paper is on the test loss, and we wish to understand how many samples can be deleted from a learnt model while still ensuring that the updated model performs well on unseen examples (i.e., the \newest{generalization error}). As we discuss in \pref{sec:pop-vs-erm}, the goal of minimizing the training loss is qualitatively different from that of minimizing the test loss.

Second, the prior work focuses exclusively on the computational cost of unlearning, without concern for associated storage requirements. This has led to approaches which involve memory-intensive checkpointing data structures, which enables fast processing of deletion requests, but consumes potentially impractical amounts of storage. In contrast, we are additionally concerned with memory usage, which highlights the drawbacks of such approaches. Unlike prior work, our algorithms do not require the training data to be available to the unlearning algorithm when deleting samples, and only rely on some cheap-to-store data statistics. 

The most closely related work to ours is the \textit{certified data removal framework} of  \cite{GuoGHV20} which provides efficient data deletion algorithms for generalized linear models (linear and logistic regression). While our deletion algorithm is similar to the Newton update removal mechanism considered in their work, there are some important technical differences. First, their unlearning setup requires access to the entire training dataset for deleting samples; we do not require this. Second, they provide theoretical guarantees in terms of the norm of the empirical gradient being small after data removal. In comparison, our guarantees are for the test loss. Third,  their unlearning definition requires the learning algorithm to be randomized, and this leads to worse performance guarantees due to added noise. In comparison, we do not need to randomize the learning algorithm. Finally, our guarantees hold for arbitrary convex loss functions and are thus broader in scope. 

Several other models of unlearning have been considered. \citet{GargGV20} give an alternative
perspective on machine unlearning, grounded in cryptography. Other works in this space focus on exploring privacy
risks~\citep{ChenZWBHZ20} and verification~\citep{sommer2020towards} in
machine unlearning settings. For specific learning models like SVMs, exact unlearning has been considered under the framework of decremental learning \citep{cauwenberghs2001incremental, tveit2003incremental, karasuyama2010multiple, romero2007incremental}. However, the primary motivation in these works is to use the framework of decremental learning to estimate the leave-one-out error in order to provide generalization guarantees for the learnt model.  
Finally, there has also been recent empirical and theoretical work in developing definitions and algorithms for machine unlearning with deep neural networks for application domains in computer vision  \citep{du2019lifelong, golatkar2020eternal, golatkar2020mixed, nguyen2020variational}.

\section{Preliminaries} 
\label{sec:definitions}

Let $\cD$ be a distribution over an instance space $\cZ$ and $\cW
\subseteq \mathbb{R}^d$ be the parameter space of a hypothesis
class.  Let $f \colon \cW \times \cZ \to \mathbb{R}$ be a loss function.  The
goal is to minimize the test loss population risk (test loss), given by
\begin{align} 
   F(w) \ldef{} \En_{z \sim \cD} [f(w, z)],
\end{align}
 where $f(w, z)$ is the loss of the hypothesis corresponding to $w\in\cW$ on the instance $z \in \cZ$.  Let $F^* = 
 \min_{w \in \cW} F(w)$ be the value of this minimum and $w^*$ be a corresponding  
 minimizer.  Since the distribution $\cD$ is \newer{often} unknown, we  are restricted to rely on samples to find a small test loss model. Given $S = (z_1, z_2, \ldots, z_n)$, a set of $n$
 samples drawn independently from $\cD$, standard learning algorithms minimize the
 empirical loss given by 
\begin{align} 
   \widehat{F}_n(w) \ldef{} \frac{1}{n} \sum_{i=1}^n f(w, z_i). \label{eq:emp_loss} 
\end{align} 

\subsection{Learning} Let $A \colon \cZ^n \to \cW$ be a learning algorithm that
takes the dataset $S$ and returns a \newer{hypothesis} $A(S) \in \cW$. The quality of $A$ is measured in terms of the difference between the 
population risk of the hypothesis $A(S)$ and the risk of the best
hypothesis $w^*$ in $\cW$, i.e., the excess risk 
\[
  \En \brk*{F(A(S))} - F^*,  
\] 
where the expectation is over the randomness in $A$ and $S$. This gives a natural notion of sample complexity. 
\begin{definition}[Sample complexity of learning]\label{def:sc-unlearning}
The $\gamma$-sample complexity of a problem is defined as 
\[ 
 n_{\gamma} := \min\{n \mid \exists A {~\rm s. t.~} \En_{S \sim \cD^n}[F(A(S))]   
 - F^*\le \gamma  \text{\quad for all ~} \cD\},   
\] 
the fewest number of samples with which a $\gamma$-suboptimal minimizer of the population loss $F(w)$ can be achieved for any distribution over the data samples. 
\end{definition}

For comparing different algorithms throughout the paper, we set $\gamma =0.01$ (or any other small arbitrary constant), and require that the provided learning algorithms guarantee an excess risk bound of $0.01$. Standard results in learning theory~\citep[Theorem
  6.1]{bubeck2014convex} show that for convex and strongly convex
losses, 
\begin{equation}
  \label{eq:standard} 
n_{0.01} = O(1), 
\end{equation} 
where the hidden constant depends on the properties of $f$ such as its Lipschitzness, but is independent of the dimension $d$ of the parameter space $\cW \subseteq \bbR^d$.  

\subsection{Unlearning} Suppose a learning algorithm $A$ over $S$ outputs the model $A(S)$. An 
unlearning algorithm $\bA$ takes as input the model  $A(S)$ and a set $U\subset S$ of 
data samples that are to be deleted, and is required to output a new model $\wt w \in \cW$. Besides the set $U$ and the model $A(S)$, the unlearning algorithm $\bar{A}$ can also access some additional data statistics $T(S)\in\cT$ (but not $S$ directly).  

This set of statistics $T(S)$ captures the additional storage required by the algorithm to support unlearning.
Thus, one of our goals is to minimize $|T(S)|$, in particular aiming for memory requirements which are independent of the training data size $n$.
This precludes strategies which involve storing and reusing the entire training set, or aggressive model checkpointing.
On the other hand, it permits storage of simple statistics such as the empirical mean, variance or average gradient of training data points, which may prove useful when unlearning. 
At the same time, we are still concerned with our unlearning algorithm's time complexity.
This goes hand in hand with the storage complexity: for most natural algorithms, the two are likely to be polynomially related. 

Augmented by this set of data statistics $T(S)$, an unlearning algorithm is a mapping $\bar{A} \colon \cZ^m
\times \cW \times \cT \to \cW$. We now define a notion of unlearning, which is motivated by the definition of differential privacy~\citep{DworkMNS06}. 

\begin{definition}[$(\varepsilon,\delta)$-unlearning] \label{def:unlearning}  For all $S$ of size $n$ and 
 delete requests $U \subseteq S$ such that $|U| \leq m$, and $W \subseteq \cW$, a learning algorithm $A$ and an unlearning algorithm  $\bar{A}$ is $(\varepsilon,\delta)$-unlearning if 
\begin{align*} 
 & \Pr \prn*{\bA(U, A(S), T(S)) \in W} \leq e^{\varepsilon} \cdot \Pr
  \prn*{\bA(\emptyset, A(S \setminus U), T(S \setminus U)) \in W} +
  \delta,
\intertext{and} 
    & \Pr \prn*{\bA(\emptyset, A(S \setminus U), T(S \setminus U)) \in W
    } \leq e^{\varepsilon} \cdot \Pr \prn*{\bA(U, A(S), T(S)) \in W} +
  \delta, 
\end{align*} where $\emptyset$ denotes the empty set and $T(S)$ denotes the data statistics available to $\bar{A}$.    
\end{definition} 

The above states that with high probability, an observer cannot
differentiate between the two cases: (i) the model is trained on the
set $S$ and then a set $U$ of $m$ points are deleted by the unlearning algorithm  using statistics $T(S)$ and (ii) the 
model is trained on the set $S \setminus U$ and no points are deleted thereafter by the unlearning algorithm. For simplicity, throughout the paper, we assume that $\epsilon \leq 1$. 

While being similar in spirit, the above notion of unlearning is different from the one considered in \citet{GinartGVZ19}. Specifically, their definition compares the output of the unlearning algorithm after deleting $m$ samples, to the output of the learning algorithm that only operates on $S \setminus U$. Thus, they require the learning algorithm to be randomized, even in the situations when there would be no delete requests in the future. Thus, the output of the learning algorithm will suffer a degradation in its performance guarantees due to this added noise. On the other hand, our definition does not require the learning algorithm to be randomized. In fact, our definition is more general than that of \citet{GinartGVZ19}. Specifically, we can simulate their comparison in our definition by considering the unlearning algorithms for which $\bar{A}$ simply adds noise to the output of $A(S \setminus U)$ when $U = \emptyset$. 

Our definition of unlearning leads to the following natural definition of the deletion capacity that formalizes how many samples can be deleted while still ensuring good test loss guarantees. 
\begin{definition}[Deletion capacity]\label{def:delection-capacity} Let $\varepsilon, \delta \geq 0$. Let $S$ be a dataset of size $n$ drawn i.i.d.\ from $\cD$, and let $f(w, z)$ be a loss function. For a pair of learning and unlearning algorithms $A, \bar{A}$ that are $(\epsilon, \delta)$-unlearning, the deletion capacity $m^{A, \bar{A}}_{\epsilon,
  \delta}(d, n)$ is defined as the maximum number of samples $U$ that can be unlearnt, while still ensuring an excess population risk of $0.01$. Specifically, 
\[ 
m^{A, \bar{A}}_{\epsilon,
  \delta}(d, n) \ldef{} \max\crl[\Big]{ m \mid \En\brk[\Big]{ \max_{U \subseteq S: \abs{U} \leq m} F(\bA(U,A(S),T(S))) -
 F^*} \le 0.01}, 
\]
where the expectation above is with respect to $S \sim \cD^n$ and output of the algorithms $A$ and $\bar{A}$. 
\end{definition} 

We are primarily interested in unlearning algorithms that have a high deletion capacity and for which $T(S)$ is small, and in particular does not grow with the dataset size $n$ (which can potentially be very large). 

\subsection{Unlearning via retraining from scratch} 

The most na\"ive yet natural baseline for unlearning is to simply retrain the model from scratch using the remaining data. 
That is, we let $\bA(U, A(S), T(S)) = A(S \setminus U)$. 
However, the straightforward method to implement this approach would require us to set $T(S)$ to contain the entire training dataset $S$, and thus $|T(S)| \geq n$. However, recall that, we aim to provide unlearning algorithms for which $T(S)$ is independent of $n$. 
Furthermore, retraining from scratch is computationally expensive -- merely reading all the data takes $\Omega(n - m)$ time, not accounting for the cost of actually running the algorithm. 
These drawbacks lead one to explore more efficient methods for unlearning. 

\section{Our results} \label{sec:our_results} 
Prior works consider unlearning from an optimization 
perspective, focusing on minimizing the empirical risk, and \newer{do no t
discuss the implications of unlearning on population risk/test loss.} As we show in the next section, the two can be significantly different objectives even for some of the simplest learning problems. 

\subsection{Population risk vs empirical risk}
\label{sec:pop-vs-erm} 
We first provide a simple example motivating our study of population
risk over empirical risk, quantified rigorously in 
\pref{thm:pop}. Consider the following mean estimation problem.  
Let $d = 1$, $\cZ=\mathbb{R}$, and the loss function $f(w,z) = (w-z)^2$. The empirical
risk of $n$ points $z_1, z_2,\ldots, z_n$ is minimized by the average $\frac{1}{n}
\sum^n_{i=1} z_i$. For this problem there is a simple unlearning
algorithm that minimizes empirical risk and also unlearns $U$
exactly.  We store the average of the points $\wh w =
\sum^n_{i=1} z_i / n$, and upon receiving a deletion request of a set
$U$ of $m$ samples, subtract those samples and renormalize to compute the minimizer of the empirical loss on the remaining training samples, i.e., we output   
\begin{equation}
\wt w = \frac{n}{n - m} \prn[\big]{\wh w - \frac{1}{n} \sum_{z \in U} z}. \label{eq:rerm_bad}
\end{equation}
The above update rule requires $T(S)$ to be of size $O(d)$ and completely deletes the samples $U$ satisfying the unlearning guarantee with $\varepsilon = \delta=0$. The returned solution $\wt w$ is the exact minimizer of the empirical loss on left over data points. 

However, $\wt w$ may not perform well on fresh samples drawn from the test
distribution.  Consider the same setting as above, but where the
points $z_i$ are drawn i.i.d.\ from $\text{Bernoulli(1/2)}$.  
Thus, the optimal parameter $w^*$ that minimizes the test loss is given by $1/2$.  However, consider the scenario where all of the $m$ delete requests are adversarially chosen and correspond to points with value $1$. In this case, the minimizer of the updated empirical loss would be smaller by an additive factor of $m/n$ than the previous estimate, and would thus have worse test loss. 

Furthermore, we note that if we only care about minimizing the empirical loss, then we can delete up to $n-1$ sample points using the procedure in \pref{eq:rerm_bad}. This applies to prior work \citep{neel2020descent, bourtoule2019machine, ullah2021machine}. However, the deletion capacity is inherently limited if we want to retain test loss guarantees. We formalize this intuition in
\pref{thm:pop} and show that even if the unlearning algorithm has 
access to all the undeleted samples $S \setminus U$, there is a
non-trivial limit on the deletion capacity. 

 \begin{theorem}\label{thm:pop}
 Let $\delta \leq 0.005$ and $\epsilon = 1$. There exists a 
    $4$-Lipschitz and $1$-strongly convex loss function $f$, and a distribution $\cD$, such that for any  learning algorithm $A$ and unlearning algorithm $\bar{A}$ which
    even has access to undeleted samples $S\setminus U$, the deletion capacity 
\[ 
m^{A, \bar{A}}_{\epsilon, \delta}(d, n) \leq cn, 
\]
where $c$ depends on the properties of
function $f$ and is strictly less than 1. 
\end{theorem} 

Finally, we remark that prior work \citep{shalev2009stochastic, feldman2016generalization} shows that there are learning settings for which  even the empirical risk minimizer solution fails to generalize. Our situation is worse, since the delete requests $U$ can be adversarially chosen from $S$ \citep{lai2016agnostic, diakonikolas2019robust}. In fact, the proof of \pref{thm:pop} relies on showing the existence of an adversary that can change the empirical loss considerably by deleting samples. We defer full details to \pref{app:pop}.

\subsection{Strict separation between unlearning and differential privacy} 
Given the strong resemblance between differential privacy and our definition for unlearning, a natural approach would be to use tools from differential privacy (DP) for machine unlearning.  The simplest
way is to ignore the particular set of delete requests $U$ and construct
an unlearning algorithm $\bar{A}$ that only depends on the learning algorithm $A(S)$. More formally, such an unlearning algorithm is of the form $\bar{A}(U, A(S), T(S)) = \bar{A}(A(S))$ and satisfies:   \begin{align*}
  \Pr \prn*{\bA(A(S)) \in W} \leq e^{\varepsilon} {\Pr \prn*{\bA(A(S 
      \setminus U)) \in W}}  + \delta, \\ \Pr \prn*{\bA(A(S\setminus
    U)) \in W} \leq e^{\varepsilon} {\Pr \prn*{\bA(A(S )) \in W}} +
  \delta.
  \end{align*} 
Note that any such pair of algorithms would be differentially private with
  respect to the original dataset $S$, where the notion of
  neighboring datasets is for datasets with edit distance of $m$.  The above 
  guarantee is stronger than the distribution-free unlearning guarantee in \pref{def:unlearning}, and thus it suffices to satisfy it. The key observation is that any DP algorithm $A$, which is private for datasets with edit distance $m$, automatically unlearns any $m$ data samples. Thus, the standard performance guarantees for DP learning yields the following bound on deletion capacity: 

  \begin{lemma}[Unlearning via DP] 
    \label{lem:unlearning_dp}
    There exists a polynomial time learning algorithm $A$ and
    unlearning algorithm $\bar{A}$ of the form $\bar{A}(U, A(S), T(S)) = A(S)$ such that the deletion capacity 
    \begin{align*}
    m^{A, \bar{A}}_{\varepsilon ,\delta} (d, n) \geq \wt \Omega \prn[\Big]{\frac{n \epsilon}{\sqrt{d \log(e^{\epsilon}/\delta)}}}, \numberthis \label{eq:del_capacity}  
\end{align*} 
where the constant in the $\Omega$-notation above depends on the properties of the loss function $f$. 
    \end{lemma} 
 The above result raises an immediate question of whether this particular
  dependence on $d$ and $n$ is necessary on the deletion capacity, and if it can be improved further. The following lemma shows that there exist
  problem instances for which any unlearning algorithm that ignores the samples
  $U$, can not improve over the factor of $\sqrt{d}$ in the denominator of the deletion capacity bound in \pref{eq:del_capacity}.
  
 \begin{lemma}[\cite{bassily2019private}, Section C] 
    \label{lem:unlearning_dp_lower}
    For any learning algorithm $A$ and an unlearning algorithm 
    $\bar{A}$ that does not use $U$, i.e., $\bar{A}(U,A(S)) =
    \bar{A}(A(S))$, there exists a $1$-strongly convex function and $O(1)$-Lipschitz loss function $f$,
    and a distribution $\cD$ such that we can not unlearn even a single sample point, if $$n \leq c \cdot {\frac{\sqrt{d}}{\epsilon}},$$ 
    where $c$ depends on the properties of the function $f$.
    \end{lemma} 

  Given that the $\sqrt{d}$ dependence on dimension is unavoidable for
  algorithms that directly use DP, it is natural to wonder whether
  this factor may be bypassed using other techniques. Our main contribution in this work a positive answer in this direction. As we show in the next section, when the loss function is convex, there exist unlearning algorithms which can delete up to $n / d^{1/4}$ sample points while still retaining the performance guarantee with respect to the test loss. 

\subsection{Unlearning for convex loss functions}  
In this section, we provide an unlearning algorithm $\bar{A}$ for convex losses that can delete more points than unlearning algorithms that simply use DP.    \begin{theorem} \label{thm:main}
    There exists a learning algorithm $A$ and an unlearning algorithm
    $\bar{A}$ such that for any convex (and hence strongly convex), L-Lipschitz, and M-\newer{Hessian-Lipschitz}
    loss $f$ and distribution $\cD$,
    \[
m^{A, \bar{A}}_{\epsilon, \delta}(d, n) \geq c \cdot \frac{n \sqrt{\epsilon}}{\prn*{d\log(1/\delta)}^{1/4}}, 
    \]
where the constant $c$ depends on the Lipschitz constants $L$ and $M$. Furthermore, for the unlearning algorithm $\bar{A}$ has running time $O(d^\omega)$ where $\omega \in [2, 2.38]$ is the exponent of matrix multiplication, and space complexity for $T(S) = O(d^2)$. \end{theorem} 
  
\newer{\pref{thm:main} and \pref{lem:unlearning_dp_lower} together show
  that  there exist problem settings, where the deletion capacity in unlearning and DP is different by a multiplicative $\Theta(d^{1/4})$. In particular,  when learning with convex loss functions, we can delete $O(n/d^{1/4})$ samples while still retaining good performance on the unseen test loss, whereas DP only guarantees deletion of $\Theta(n/d^{1/2})$ samples. Hence, our algorithm is at least quadratically better in terms of dependence on $d$ in deletion capacity that standard DP algorithms.} Besides improving the dependence on $d$, our algorithm also enjoys better dependence on $\epsilon$ and $\log(1/\delta)$ in the deletion capacity, by at least a quadratic factor.
  
Our learning algorithm stores additional statistics of the dataset $S$ in order to delete the set $U$ in the unlearning algorithm. The extra memory we need for these statistics is independent of $n$. Furthermore, our algorithm uses the samples in $U$ during the unlearning phase. This paradox of storing and using information in order to delete it, motivates the name of the paper: \textit{Remember what you want to forget}.   
   
  Characterizing the entire set of problems for
  which unlearning and differential privacy are different remains an 
  interesting open question. \pref{thm:main} yields an improved upper bound, 
  but it is not clear if this dependence on $d$ or $n$ in the deletion capacity is tight or if it can be improved even further. Resolving this question would be a fascinating future research direction.

\section{Unlearning algorithms}   \label{sec:sc_algorithms}

In the following, we provide learning and unlearning algorithms when
the loss function $f(\cdot, z)$ is $\lambda$-strongly convex. The unlearning algorithms for convex losses follows by appealing to the strongly convex case after adding regularization. We defer the algorithms and proofs for the convex case to \pref{app:convex}. Throughout this section, we make the following assumption: 
\begin{assumption}
\label{ass:f_properties} 
For any $z \in \cZ$, the function $f(w, z)$ is $\lambda$-strongly convex, $L$-Lipschitz
and $M$-Hessian Lipschitz with respect to $w$. 
\end{assumption} 

\paragraph{Learning algorithm.}  We denote our learning algorithm by $A_{sc}$. When given a dataset $S$ of $n$ points sampled i.i.d.\ from some distribution 
$\cD$, the algorithm $A_{sc}$ computes the point $\wh w$ by  minimizing the empirical loss $\wh F_n(w)$, i.e.  
\begin{align*}
\wh w \leftarrow \argmin\, \wh F_n(w) \ldef{} \frac{1}{n}  \sum_{z \in S} f(w, z). \numberthis \label{eq:erm_alg}
\end{align*}
$A_{sc}$ then returns the point $\wh w$ and the statistics $T(S) \ldef{} \crl{ \grad^2 \wh F(\wh w)}$ containing the Hessian of $\wh F(w)$ evaluated at the output point $\wh w$. We provide the pseudocode for $A_{sc}$ in the appendix. 

\paragraph{Unlearning algorithm.} We denote our unlearning algorithm by $\bar{A}_{sc}$ and provide the pseudocode in \pref{alg:unlearningalg_rerm_sco}. $\bar{A}_{sc}$ receives as input the set of
delete requests $U$, the point $\wh w$ and the data statistics $T(S)$. 
Using these inputs,  $\bar{A}_{sc}$ first estimates the
matrix $\wh H$ that denotes the Hessian of the empirical function on
the dataset $S \setminus U$ when evaluated at the point $\wh w$. Then, $\bar{A}_{sc}$ computes the point
$\bar{w}$ by removing the contribution of the deleted points $U$ from
$\wh w$ using the update in \pref{eq:bar_def}. Finally, $\bar{A}_{sc}$ perturbs $\bar{w}$ with noise $\nu$ drawn from $\cN(0, \sigma^2 \bbI_d)$ and returns the perturbed point $\wt w$. 

\begin{algorithm}
\caption{Unlearning algorithm ($\bar{A}_{sc}$)} 
\begin{algorithmic}[1]
   \Require Delete requests: $U = \crl*{z_j}_{j=1}^{m} \subseteq S$, output of $A_{sc}(S)$: $\wh w$, additional statistic $T(S): \crl{\grad^2 \wh F(\wh w)}$, loss function: $f(w, z)$. 
   \State Set $\gamma = \frac{2Mm^2 L^2}{\lambda^3 n^2}$, $\sigma = \frac{\gamma}{\epsilon} \sqrt{2 \ln(1.25 / \delta)}.$ 
   \State\label{line:gradient_estimation}Compute 
   \begin{align*}
   \wh H = \frac{1}{n-m} \prn[\big]{n \grad^2 \wh F(\wh w) - \sum_{z \in U} \grad^2 f(\wh w, z)}. \numberthis \label{eq:H_hat_def} 
   \end{align*} 
   
   \State\label{line:bar_def}Define 
\begin{align*}
\bar{w} = \wh w +  \frac{1}{n - m} \prn{\wh H}^{-1} \sum_{z \in U} \grad f(\wh w, u). \numberthis \label{eq:bar_def}
\end{align*}   
   \State Sample $\nu \in \bbR^d$ from $\cN(0, \sigma^2 \bbI_d)$.   
   \State \textbf{Return} $\wt w \ldef \bar{w}  +  \nu$.  \end{algorithmic}  
\label{alg:unlearningalg_rerm_sco}  
\end{algorithm}

Our main technical insight that leads to improvements in deletion capacity over differential privacy is the following observation. For loss functions that satisfy \pref{ass:f_properties}, when deleting $m$ samples, we can approximate the empirical minimizer on the dataset $S \setminus U$ up to a precision of $O(m^2 / n^2)$ by the point $\bar{w}$ computed in \pref{eq:bar_def}. This implies that we only need to add noise of the scale of $\sigma \propto O(m^2 / n^2)$ to get the desired unlearning guarantee. This noise is smaller than the amount of noise typically added for DP learning \citep{DworkR14} by a quadratic factor; hence giving us a quadratic improvement in the deletion capacity. 
\begin{lemma} 
\label{lem:w_update_v_main} 
Suppose the loss function $f$ satisfies \pref{ass:f_properties}. Let $S \sim \cD^n$ be a set of $n$ samples, and $U \subseteq S$ denote the set of $m$ delete requests. Define the point $\wh w'$ as the empirical minimizer over $S \setminus U$, i.e. $  {\wh w'} \in  \argmin_{w} \sum_{z \in S \setminus U} f(w, z) / (n - m)$. Then,  
$$\nrm{\wh w' - \bar{w}} \leq \frac{2 M L^2m^2}{\lambda^3 n^2}, $$
where the point $\bar{w}$ is defined in \pref{eq:bar_def} in \pref{alg:unlearningalg_rerm_sco}. 
\end{lemma} 

The following theorem provides performance guarantees for the algorithms  $A_{sc}$ and $\bar{A}_{sc}$, and show that $A_{sc}$ and $\bar{A}_{sc}$ are $(\epsilon, \delta)$-unlearning. 

\begin{theorem} Suppose the loss function $f$ satisfy  \pref{ass:f_properties} and let the dataset $S \sim \cD^n$. Then, 
\label{thm:sc_main_thm}   

\begin{enumerate}[label=$(\alph*)$] 
\item The point $\wh w$ returned by running $A_{sc}$ on $S$ satisfies 
\begin{align*}  
	\En_{S \sim \cD^{n}} \brk{ F(\wh w) - \min_{w \in \cW} F(w)} \leq \frac{4 L^2}{\lambda n}.  \numberthis \label{eq:bound1} 
\end{align*} 

\item For any set $U \subseteq S$ of $m$ delete requests,  the point $\wt w$ returned by $\bar{A}_{sc}$ satisfies 
\begin{align*}
	\En_{S, \nu} \brk{F(\wt w) - \min_{w \in \cW} F(w)} = O\prn[\Big]{\frac{ \sqrt{d} M m^2 L^3}{\lambda^3 n^2 \epsilon} \sqrt{\ln\prn[\big]{1/\delta}} + \frac{4mL^2}{\lambda n}}. \numberthis \label{eq:bound2} 
\end{align*} 
 
\item The learning algorithm $A_{sc}$ and the unlearning algorithm $\bar{A}_{sc}$ are $(\epsilon, \delta)$-unlearning. 
\end{enumerate}
\end{theorem} 

We defer the proof to the \pref{app:sc_main_thm}. The above excess risk guarantees for the output of algorithms $A_{sc}$ and $\bar{A}_{sc}$ give a lower bound on the number of samples $m$ that can be deleted while still ensuring the desired excess risk guarantee (deletion capacity). Specifically, from the performance guarantee for $\bar{A}_{sc}$ in \pref{eq:bound2}, we observe that we can delete 
    \[ 
m \geq c \cdot \frac{n \sqrt{\epsilon}}{\prn*{d\log(1/\delta)}^{1/4}}, 
    \] 
  samples from the set $S$ (with size $n$) while still ensuring that an excess risk guarantee of $\gamma = 0.01$. Here,  $c$ depends on the constants $M, K$ and $\lambda$ for the function $f$. This proves \pref{thm:main} for strongly convex loss functions. 
  
\paragraph{Memory.}
We do not need to store the entire dataset $S$ for the unlearning algorithm $\bar{A}_{sc}$. We note that the data statistic $T(S)$ that is passed as an input to $\bar{A}_{sc}$ is given by 
$T(S) = \crl{\grad^2 \wh F(\wh w)}$.  Clearly, $T(S)$ needs $O(d^2)$ memory and thus $\abs{T(S)}$ is independent of $n$ or $m$. 

\paragraph{Computation.} For the sake of exposition above, our learning algorithm $A_{sc}$ computes the exact minimizer for the empirical loss  in \pref{eq:erm_alg}. However, as we show in \pref{app:sc_main_thm}, our theoretical guarantees hold even when the empirical minimizer is computed approximately up to a precision of $O(1/n^2)$.  When the domain $\cW$ is convex, such a minimizer can be efficiently computed using standard optimization algorithms like accelerated gradient descent, SGD, clipped-SGD, etc. For example, for the $\lambda$-strongly convex case, Nesterov's accelerated GD algorithm can find a $O(1/n^2)$ approximate minimizer in time $\wt O(n d / \sqrt{\lambda})$ \citep{bubeck2014convex, nemirovskij1983problem}. Furthermore, $A_{sc}$ takes $O(n d^2)$ time to compute $T(S)$. 

On the other hand, the running time for the unlearning algorithm
$\bar{A}_{sc}$ scales as $O(d^\omega)$, the time taken to
invert the matrix $\wh H$. Here, $\omega \in [2, 2.38]$. Note that our
unlearning time is independent of the (potentially large) dataset size
$n$. \newest{Furthermore, for problems
such as linear SVMs where the Hessian is a diagonal matrix, $A_{sc}$ takes
time $O(nd)$ and $\bar{A}_{sc}$ takes time
$O(d)$.}

\paragraph{Algorithms for convex losses.}  Our unlearning algorithms when the loss function is convex are based on reductions to the strongly convex setting discussed above. Give the convex loss function $f(\cdot, z)$, we define the function $\wt f(\cdot, z)$ as 
\begin{align*} 
\wt f(w, z) = f(w, z) + \frac{\lambda}{2} \nrm*{w}^2. 
\end{align*} 
The key observation is that the function $\wt f(w, z)$ is $\lambda$-strongly convex, $(L + \lambda \nrm{w})$-Lipschitz, $(H + \lambda)$-smooth and $M$-Hessian Lipschitz, and thus we can run algorithms $A_{sc}$ and $\bar{A}_{sc}$ on $\wt f$. We defer the algorithmic implementation and  theoretical analysis for convex loss setting to \pref{app:convex}.

\section{Conclusion} 
We initiated a new study on machine unlearning with a focus on population risk minimization, in comparison to previous works that focus on empirical risk minimization. For the case of convex loss functions, we provide a new unlearning algorithm that improves over the deletion capacity, by at least a quadratic factor in $d$, than using an out of the box differentially private algorithm for unlearning.  Proving a dimension dependent information theoretic lower bound on the deletion capacity is an interesting future research direction. Another exciting direction of future research is to provide efficient unlearning algorithms for finite / discrete hypothesis class, and for non-convex loss functions. Finally, in this work, we considered the problem of batch deletion where the delete request $U$ all arrive at the same time; Extending our algorithms for the online case is another  interesting research direction that we are excited to pursue in future research. 

\subsection*{Acknowledgements} 
We thank Robert Kleinberg, Mehryar Mohri, and Karthik Sridharan for helpful discussions. JA is supported in part by the grant NSF-CCF-1846300 (CAREER), NSF-CCF-1815893, and a Google Faculty Fellowship. GK is supported by an NSERC Discovery Grant. 

\clearpage 

\setlength{\bibsep}{6pt}
\bibliography{refs,biblio} \bibliographystyle{icml2021}
\clearpage

\newpage
\appendix 
\renewcommand{\contentsname}{Contents of Appendix}
\tableofcontents  
\addtocontents{toc}{\protect\setcounter{tocdepth}{3}} 
\clearpage 

\setlength\parindent{0pt}
\setlength{\parskip}{0.25em} 

\section{Additional Notation}  
We recall the following standard definitions for the loss function $f(w; z)$. 
\begin{definition}[Lipschitzness] 
The function $f(w, z)$ is $L$-Lipschitz in the parameter $w$ if  for all $z \in \cZ$, and all $w_1, w_2 \in \cW$, 
$$|f(w_1,z) - 
f(w_2,z)| \leq L \|w_1 - w_2\|.$$ 
\end{definition} 

\begin{definition}[Strong convexity] The function $f(w, z)$ is $\lambda$-strongly convex, if  for all $z \in \cZ$, and all $w_1, w_2 \in \cW$,
\[
f(w_1, z) \geq f(w_2, z) + \tri{\nabla f(w_2, z), w_1 - w_2} +
\frac{\lambda}{2} \|w_1 -w_2\|^2.
\] 
\end{definition} 

\begin{definition}[Hessian-Lipschitzness] The function $f(w, z)$ is said to be $M$-Hessian Lipschitz if  for all $z \in \cZ$, and all $w_1, w_2 \in \cW$,\begin{align*}
\nrm{\grad^2 f(w_1, z) - \grad^2 f(w_2, z)} \leq M \nrm*{w_1 - w_2},  
\end{align*}
or equivalently, that $\nrm{\grad^3 f(w, z)} \leq M$ for all $w$.    
\end{definition} 

\section{Missing proofs from \pref{sec:our_results}}  
\subsection{Proof of \pref{thm:pop}} \label{app:pop} 
We first develop some technical results, which we will use to prove
\pref{thm:pop}. For a distribution $\cD$, define $\mu(\cD) \ldef{} \En_{Z \sim
  \cD} [Z]$.

\begin{lemma}
  \label{lem:hyp}
  There exists two distributions $\cD_1$ and $\cD_2$ over
  $[0,1]$ such that $\|\cD_1 - \cD_2\|_1 \leq \alpha$ and
  $|\mu(\cD_1) - \mu(\cD_2)| \geq \alpha/2$.
\end{lemma}
\begin{proof}
Let $\cD_1$ be the uniform distribution over $[0.25 + \alpha/2, 0.75 +
  \alpha/2]$ and $\cD_2$ be the uniform distribution over $[0.25,
  0.75]$. We note that $\En_{z \sim \cD_1}  \brk*{z} = 0.5 + \alpha/2$ and $\En_{z \sim \cD_2}  \brk*{z} = 0.5$ and hence
$|\En_{z \sim \cD_1}  \brk*{z} - \En_{z \sim \cD_2}  \brk*{z}| =\alpha/2$. However, the $\ell_1$ distance between
$\cD_1$ and $\cD_2$ is bounded by $\alpha$. 
\end{proof} 

\begin{lemma}
  \label{lem:adv}
Given two distributions $\cD_1$ and $\cD_2$ over the domain $\cZ$, let the distribution $\bar{\cD}$ be defined such that for any $z \in \cZ$, $\bar{\cD}(z) \propto \min(\cD_1(z), \cD_2(z))$. Then, for any $\cD_1$ and $\cD_2$ such that $\nrm{\cD_1 - 
  \cD_2}_1 =  \alpha$, and $n$ samples $\crl{z_i}_{i=1}^n$ drawn iid from $\cD_1$, there
  exists an adversary that deletes at most $2 n \alpha$ samples and
  outputs the dataset $\crl{z_j}_{j=1}^{n'}$ with a distribution $\widetilde{\cD}^{n'}$ and $n' \leq
  n - 2 n \alpha$ such that
  \[
\| \wt{\cD}^{*} - \bar{\cD}^{*}\|_1 \leq e^{-  n \alpha /6},
  \] 
  where the superscript denotes distribution over all sequences over the domain $\cZ$.  
  \end{lemma} 
\begin{proof}
  Let $m = 2 n \alpha$. Our proof uses two adversaries $A'$ and $A$. We first define $A'$. Given $n$ samples $\crl{z_i}_{i=1}^n$, it deletes sample
  $z_i$ with probability $\frac{\brk{\cD_1(z_i) -
    \cD_2(z_i)}_+} {\cD_1(z_i)}$, where $\brk{x}_+ \ldef{} \max\crl{x, 0}$. First
  observe that for any sample $z$, probability  that $z$ is retained is given by 
  \[
  \cD_1(z) \cdot \frac{\min\crl{\cD_1(z), \cD_2(z)}}{\cD_1(z)} =
  \min\crl{\cD_1(z), \cD_2(z)}.
  \]
  Thus the distribution of samples outputted by $A'$ is exactly
  $\bar{\cD}$. However, it can delete more than $m$ samples.  Next,  we consider another adversary $A$, which is same as $A'$, except it
  stops after deleting $m$ samples.  Hence, for sequences of length
  $> n - m$, the output of $\cA'$ and $\cA$ are the same.  Hence,
  \[
\| \widetilde{\cD}^{*} - \bar{\cD}^{*}\|_1 \leq \Pr(N \geq m), \numberthis \label{eq:temp1}
\]
where $N$ is the number of deleted samples. In the rest of the proof,
we bound this probability.

Let $X_i \in \crl{0, 1}$ be the random variable that takes the value $1$ if the sample $i$ is deleted. Hence $N =
\sum_i X_i$. Furthermore, $\En[X_i] = \int_z \brk{\cD_1(z), \cD_2(z)}_+ dz =
\|\cD_1 - \cD_2\|_2 / 2 =  m / 4n$.
By the Chernoff bound, we have that 
\begin{align*} 
\Pr\prn[\Big]{\sum_i X_i \geq m} 
 \leq e^{-m/12}.
\end{align*}
Using the above with \pref{eq:temp1} implies the desired statement. 
\end{proof}

We now have all the tools to prove \pref{thm:pop}.
\begin{proof}[Proof of \pref{thm:pop}] 
The proof for small values of $m$ follows from known bounds for sample complexity of learning \citep{bubeck2014convex, shalev2014understanding}. In the following, we provide an information-theoretic lower bound for $m \geq 100$ by
constructing two distributions $\cD_1$ and $\cD_2$ and show that no
single learning-unlearning pair $A, \bar{A}$ can perform well on both
of them.

Let $\cW = [0,1]$ and $\cZ = R$. Further, let the loss function be $f(w,z) = (w - z)^2$. Our
proof consists of two main parts: we first provide a reduction from learning to mean
estimation, and then give the lower bound by a reduction from mean estimation to
hypothesis testing.

\paragraph{Reduction from learning to mean estimation:} We first show
that for any distribution $\cD$ for the population loss given by $F(w) \ldef{} \En_{z \sim D} \brk*{f(w z)}$ satisfies, 
\[
F(w) - F(w^*) = (w-w^*)^2.
\] 
To observe this note that
\begin{align*}
  F(w) & = \En[(w-z)^2]  \\
  & = \En[(w-w^* + w^* - z)^2]  \\
  & = \En[(w-w^*)^2] + 2  \En[(w-w^*)(w^* - z)] + \En[(w^*-z)^2] \\
  & = (w-w^*)^2 + 2(w-w^*)\En[(w^*-z)] + F(w^*) \\
  & = (w-w^*)^2 + F(w^*),
  \end{align*}
where the last inequality uses the fact that $\En[z]= w^*$ for our loss function. Thus, in order to bound the learning error, it suffices to bound the error in
estimating the mean of the underlying distribution $w^*$.

\paragraph{From mean estimation to the lower bound:} 
Since $\cZ$ is unbounded and having
more information only helps, we assume that the unlearner has access
to the entire sample set $S$ i.e., the passed data statistics $T(S) = \crl{S}$. Since the output $A(S)$ can be
derived from $S$, it suffices to consider unlearning algorithms
$\bar{A}$ of the form $\bar{A}(U,S)$.  Let $w$ denote the output $\bar{A}(U,S)$. By the
definition of forgetting rule,
\begin{align*}
  \En_{\cD_1} (\bar{A}(U, A(S)) -w^*_1)^2 & = \int_{w}
  \Pr_{\cD_1}(\bar{A}(U, A(S)) = w) (w -w^*_1)^2 dw \\ & \geq
  e^{-\epsilon} \int_{w} \Pr_{\cD_1}(\bar{A}(\phi, A(S \setminus U)) =
  w) (w -w^*_1)^2 dw - \delta, 
  \end{align*}
where the last inequality uses the fact that $\cW = [0,1]$.  Let 
$\tilde{A} = \bar{A}(\phi, A(\cdot))$. Let $\alpha = m / 2n$ and let
$\cD_1$ and $\cD_2$ be two distributions such that $\|\cD_1 -
\cD_2\|_1 = \alpha$.  Let the set of delete requests $U$ be chosen by the
adversary in Lemma~\ref{lem:adv}. For these set of samples $U$, we have that 
\begin{align*}
 \int_{w} \Pr_{\cD_1}(\bar{A}(\phi, A(S \setminus U)) = w) (w
  -w^*_1)^2  dw\geq  \int_{w} \Pr_{\bar{\cD}}(\bar{A}(\phi, A(S \setminus U)) = w) (w 
  -w^*_1)^2 dw - e^{-m/12},
  \end{align*}
where $\bar{\cD}$ is defined in \pref{lem:adv}.
Similarly, we have that 
\[
\En_{\cD_2} (\bar{A}(U, A(S)) -w^*_1)^2 \geq e^{-\epsilon} \int_{w}
\Pr_{\bar{\cD}}(\bar{A}(\phi, A(S \setminus U)) = w) (w -w^*_1)^2 dw -
e^{-m/12} - \delta.
\]
The sum of errors the unlearner makes on either of the errors is at least 
\begin{align*}
 \hspace{1in} & \hspace{-1in} \En_{\cD_1} [F(w)] - F^*_{\cD_1} + \En_{\cD_2} [F(w)] - F^*_{\cD_2}
  \\ & \geq e^{-\epsilon} \int_{w} (\Pr_{\bar{\cD}}(\tilde{A}(S
  \setminus U)= w) ( (w-w^*_1)^2 + (w-w^*_2)^2 ) dw - 2 \delta - 2 e^{-m/12}  \\
  & 
  \geq \frac{e^{-\epsilon}}{2} \int_{w} (\Pr_{\bar{\cD}}(\tilde{A}(S
  \setminus U)=w) ( (w^*_1-w^*_2)^2 )dw - 2 \delta - 2 e^{-m/12}  \\
  & = 
  \frac{e^{-\epsilon} (w^*_1 - w^*_2)^2}{2} - 2 e^{-m/12} - 2 \delta,  \numberthis \label{eq:error_bound1} 
\end{align*} where the inequality in the second last line follows from the fact that $a^2 + b^2 \geq (a + b)^2/2$ for any $a, b \in \bbR$. 

Plugging in the distributions $\cD_1$ and $\cD_2$ as given in \pref{lem:hyp} with $\alpha = m/2n$ in the error bound \pref{eq:error_bound1} implies that 
\begin{align*}
 \En_{\cD_1} [F(w)] - F^*_{\cD_1} + \En_{\cD_2} [F(w)] - F^*_{\cD_2} &\geq  \frac{e^{-\epsilon} m^2}{32 n^2} - 2 e^{-m/12} - 2 \delta. 
\end{align*} 
Since $\delta \leq 0.005$, $\epsilon
\leq 1$ and $m \geq 100$, for the above quantity to be less than $0.01$, we have that $m \leq c n \cdot e^{\epsilon}$, where $c e^\epsilon < 1$. 
\end{proof} 

\subsection{Proof of \pref{lem:unlearning_dp}}
  \begin{proof} We let $A$ to be a DP algorithm which is private for datasets with edit distance $m$. Our unlearning algorithm $\bar{A}$ simply returns the input point $A(S)$ without making any changes to it, i.e. $\bar{A}(U, A(S), T(S)) = A(S)$. Clearly, the unlearning algorithm $\bar{A}$ does not require any additional data statistics and thus $T(S) = \emptyset$.
  
We set the DP algorithm $A$ as the mini-batch noisy SGD method from \citet{bassily2019private}. The learning guarantee for $A$ from  \citet[Theorem 3.2]{bassily2019private} 
    together with the group privacy property of differential privacy \citep[Lemma 2.2]{vadhan2017complexity} implies that: 
 \begin{align*} 
F(A(S)) - F^* \leq 10 BL\prn[\Big]{ \frac{1}{\sqrt{n}} + \frac{m\sqrt{d \log(m e^{\epsilon} /  \delta)}}{\epsilon n}}.  \numberthis \label{eq:temp_eqn1} 
\end{align*} 
Furthermore, $A(S)$ is $(\epsilon, \delta)$-DP for datasets with edit distance $m$, i.e. for any set $U \subseteq S$ of $m$ samples: 
  \begin{align*}
 & \Pr \prn*{A(S) \in W} \leq e^{\varepsilon} {\Pr \prn*{A(S 
      \setminus U) \in W}}  + \delta, \\ &\Pr \prn*{A(S\setminus
    U) \in W} \leq e^{\varepsilon} {\Pr \prn*{A(S ) \in W}} +
  \delta.
  \end{align*} 

Since $T(S) = \emptyset$ and $\bar{A}(A(S), U, T(S)) = A(S)$ for any $U \subseteq S$ such that $\abs{U} = m$, we can rewrite the DP guarantee as: 
  \begin{align*} 
 & \Pr \prn*{\bA(U, A(S), T(S)) \in W} \leq e^{\varepsilon} \cdot \Pr
  \prn*{\bA(\emptyset, A(S \setminus U), T(S \setminus U)) \in W} +
  \delta, \\ 
    & \Pr \prn*{\bA(\emptyset, A(S \setminus U), T(S \setminus U)) \in W
    } \leq e^{\varepsilon} \cdot \Pr \prn*{\bA(U, A(S), T(S)) \in W} +
  \delta, 
\end{align*} 

implying that the pair $(A, \bar{A})$ is $(\epsilon, \delta)$-unlearning for $U$ of size $m$.

We next bound the deletion complexity. The bound in the right hand side of  \pref{eq:temp_eqn1} implies that we can delete 
\begin{align*}
m  = \wt{\Omega} \prn[\Big]{\frac{0.01 \epsilon}{\sqrt{\log(e^\epsilon / \delta)}} \cdot \frac{n}{\sqrt{d}}} 
\end{align*} samples while still ensuring that the excess risk is bounded by $\gamma = 0.01$. The above implies the desired lower bound on the deletion capacity.  
\end{proof}

\subsection{Proof of \pref{thm:main}} 
 
The following lower bound on the deletion capacity is based on the excess risk guarantees for our learning and unlearning algorithms given in  \pref{thm:sc_main_thm} and \pref{thm:convex_main_thm} (in \pref{app:convex}) for strongly convex and convex loss setting respectively. 

\begin{proof} We consider the strongly convex loss and convex loss setting separately below. 
\paragraph{Strongly convex loss setting.} Our learning algorithm $A_{sc}$ and the unlearning algorithm $\bar{A}_{sc}$ are given in   \pref{alg:learningalg_rerm_sco} and \pref{alg:unlearningalg_rerm_sco} respectively.  
\pref{thm:sc_main_thm} implies that the learning algorithm $A_{sc}$ and the unlearning algorithm $\bar{A}_{sc}$ are $(\epsilon, \delta)$-unlearning. Furthermore, we have that 
\begin{align*} 
	& \En \brk{ F(\wh w) - F^*} \leq \frac{4 L^2}{\lambda n}, 
\intertext{and}
	& \En \brk{F(\wt w) - F^*} = O\prn[\Big]{\frac{ \sqrt{d} M m^2 L^3}{\lambda^3 n^2 \epsilon} \sqrt{\ln\prn[\big]{1/\delta}} + \frac{4mL^2}{\lambda n}},
\end{align*} 
where $\wh w$ denotes the output point $A_{sc}(S)$ and $\wt w$ denotes the output point $\bar{A}_{sc}(U, A_{sc}(S), T(S))$. 

The above upper bound on the excess risk implies that we can delete at least 
    \[ 
m = c \cdot \frac{n \sqrt{\epsilon}}{\prn*{d\log(1/\delta)}^{1/4}}, 
    \]
 samples while still ensuring an excess risk guarantee of $\gamma = 0.01$. Here, the constant $c$ depends on the constants $M, L$ and $\lambda$ for the function $f$. This gives us the desired lower bound on the deletion capacity $m^{A_{sc}, \bar{A}_{sc}}_{\epsilon, \delta}(d, n)$.

\paragraph{Convex loss setting.} Our learning algorithm $A_c$ and the unlearning algorithm $\bar{A}_c$ are given in \pref{alg:learningalg_sco} and \pref{alg:unlearningalg_sco} respectively.  
\pref{lem:dp_guarantee_convex} implies that the learning algorithm $A_c$ and the unlearning algorithm $\bar{A}_{c}$ are $(\epsilon, \delta)$-unlearning. Furthermore, as a consequence of \pref{corr:corollary2}, we note that setting $\lambda$ as in \pref{eq:setting_lambda} implies that: 
\begin{align*}
\small \En \brk*{F(\wh w) - F^*} = O\prn[\Big]{c_1\sqrt{\frac{m}{n}} + c_2 \prn[\Big]{\frac{d \log(1/\delta)}{\epsilon^2}}^{1/8} \sqrt{\frac{m}{n}}} \\
\intertext{and}
\small \En \brk*{F(\wt w) - F^*} = O\prn[\Big]{c_1 \sqrt{\frac{m}{n}} + c_2 \prn[\Big]{\frac{d \log(1/\delta)}{\epsilon^2}}^{1/8} \sqrt{\frac{m}{n}}},  
\end{align*} 
where $\wh w$ denotes the output point $A_{c}(S)$ and $\wt w$ denotes the output point $\bar{A}_{c}(U, A_{c}(S), T(S))$,  and the constants $c_1$ and $c_2$ depend on the properties of the function $f$. 

The above upper bound on the excess risk implies that we can delete at least 
    \[ 
m = c \cdot \frac{n \sqrt{\epsilon}}{\prn*{d\log(1/\delta)}^{1/4}}, 
    \]
 samples while still ensuring an excess risk guarantee of $\gamma = 0.01$. Here, the constant $c$ depends on the constants $M, L$ and $B$ for the function $f$. This gives us the desired lower bound on the deletion capacity $m^{A_{c}, \bar{A}_{c}}_{\epsilon, \delta}(d, n)$. 

 \end{proof}

\section{Missing details from \pref{sec:sc_algorithms}} 

\subsection{Proof of \pref{lem:w_update_v_main}} 
\begin{lemma} 
\label{lem:w_not_too_far} The points $\wh w$ and $\wh w'$, defined in \pref{lem:w_update_v_main}, satisfy the following guarantee 
\begin{align*}
\nrm{\wh w - \wh w'} \leq \frac{2 mL}{\lambda n}. 
\end{align*}
\end{lemma}
\begin{proof} Define the functions $\wh F_1$ and $\wh F_2$ as 
\begin{align*} 
 \wh F_1(w) \ldef{} \frac{1}{n}\sum_{z \in S} f(w, z) \qquad \text{and,} \qquad {\wh F_2}(w) \ldef{}  \frac{1}{n - m} \sum_{z \in \bar{S}} f(w, z), \qquad
\end{align*} where the set $\bar{S} \ldef{} S \setminus U$. Note that $\wh w = \argmin_{w} \wh F_1(w) \text{~and,~}  \wh w' = \argmin_{w} \wh F_2(w)$. We first observe that 
\begin{align*} 
n\prn[\big]{\wh F_1(\wh w') - \wh F_1(\wh w)} &=  \sum_{z \in S} f(\wh w', z)  -  \sum_{z \in S} f(\wh w, z)  \\ 
											  &= \sum_{z \in \bar{S}} f(\wh w', z)  -  \sum_{z \in \bar{S}} f(\wh w, z) + \sum_{z \in U} f(\wh w', z)  -  \sum_{z \in U} f(\wh w, z) \\  
											  &= (m-n)\prn[\big]{\wh F_2(\wh w') - \wh F_2(\wh w)} + \sum_{z \in U} f(\wh w', z)  -  \sum_{z \in U} f(\wh w, z) \\ 
											  &\overleq{\proman{1}} \sum_{z \in U} f(\wh w', z)  -  \sum_{z \in U} f(\wh w, z) 	\overleq{\proman{2}} m L \nrm*{ \wh w' - \wh w},   \numberthis \label{eq:w_not_too_far1} 
\end{align*} where the equality in the second line above follows from the fact that $\bar{s} = S \setminus U$ and the equality in the third line holds from the definition of the function $\wh F_2$. The inequality $\proman{1}$ holds because $\wh w'$ is the minimizer of the function $\wh F_2(w)$ and the inequality $\proman{2}$ is due to the fact that the function $f$ is $L$-lipschitz. Next, note that the function $\wh F_1$ is $\lambda$-strongly convex. Thus,  
\begin{align*}
\wh F_1(\wh w') - \wh F_1(\wh w) &\geq \frac{\lambda}{2} \nrm*{\wh w - \wh w'}^2. \numberthis \label{eq:w_not_too_far2}
\end{align*} 
Using \pref{eq:w_not_too_far1} and \pref{eq:w_not_too_far2}, we get that 
\begin{align*}
 \frac{\lambda n}{2} \nrm*{\wh w - \wh w'}^2 \leq mL \nrm*{\wh w - \wh w'},
\end{align*} which implies that $\nrm*{\wh w - \wh w'} \leq \frac{2 mL}{\lambda n}.$ 
\end{proof} 

\begin{proof}[Proof of \pref{lem:w_update_v_main}] Given the function $f$ that satisfies \pref{ass:f_properties},  define the functions $\wh F_1$ and $\wh F_2$ as 
\begin{align*}
 \wh F_1(w) \ldef{} \frac{1}{n} \sum_{z \in S} f(w, z) \qquad \text{and,} \qquad {\wh F_2}(w) \ldef{}  \frac{1}{n - m} \sum_{z \in \bar{S}} f(w, z), \qquad
\end{align*} where the set $\bar{S} \ldef{} S \setminus U$. Using the Taylor's expansion for $\grad \wh F_2({\wh w'})$ around the point $\wh w$, we get that 
\begin{align*} 
\nrm{ \grad \wh F_2({\wh w'}) - \grad \wh F_2(\wh w) - \grad^2 \wh F_2(\wh w)[{\wh w'} - \hw]} \leq \frac{M}{2} \nrm{\wh w - {\wh w'}}^2, 
\end{align*} where $M$ denotes the Hessian-Lipschitz constant for the function $f(\cdot, z)$, i.e. $\nrm{\grad^3 \wh F_2(\wh w)} \leq M$. Since $\wh w'$ is a minimizer of $\wh F_2$, and $\wh F_1$ is smooth, we have that $\grad \wh F_2({\wh w'}) = 0$. Plugging this in the above bound, we get 
\begin{align*} 
\nrm{ \grad \wh F_2(\wh w)  + \grad^2 \wh F_2(\hw)[{\wh w'} - \hw]} \leq \frac{M}{2} \nrm{\wh w - {\wh w'}}.  \numberthis \label{eq:w_update_v1}
\end{align*}	
Also note that  
\begin{align*}
	\grad \wh F_2(\wh w) &=  \frac{1}{n - m} \sum_{z \in \bar{S}} \grad f(\wh w, z) \\ 
								   &= \frac{1}{n-m}  \sum_{z \in S} \grad f(\wh w, z) -  \frac{1}{n -m }\sum_{z \in U} \grad f(\wh w, z) \\
								   &= \frac{n}{n - m} \grad \wh F_1(\wh w) -  \frac{1}{n -m }\sum_{z \in U} \grad f(\wh w, z) \\ &= - \frac{1}{n -m }\sum_{z \in U} \grad f(\wh w, z)
\end{align*} where the equality in the second line above holds because $\bar{S} = S \setminus U$, the third line follows by using the definition of the function $\wh F_1(w)$, and the last line is due to the fact that $\wh w $ is the minimizer for the function $\wh F_1(w)$ and hence $\grad \wh F_1(\wh w) = 0$. Plugging the above in \pref{eq:w_update_v1}, we get that 
\begin{align*} 
\nrm{- \frac{1}{n - m} \sum_{z \in U} \grad f(\wh w; u) +  \grad^2 \wh F_2(\hw)[{\wh w'} - \hw]} \leq \frac{M}{2} \nrm{\wh w - {\wh w'}}^2.  \numberthis \label{eq:w_update_v2} 
\end{align*} 
Now, let us define the vector $v$ such that  
\begin{align*}
\wh w'  = \wh w +  \frac{1}{n - m} \prn{\grad^2 \wh F_2 (\wh w)}^{-1} \sum_{z \in u} \grad f(\wh w; z) +  v. \numberthis \label{eq:w_update_v7}  
\end{align*}
Plugging the above relation in \pref{eq:w_update_v2}, we get that 
\begin{align*} 
	\nrm{ \grad^2 \wh F_2(\hw) v} \leq \frac{M}{2} \nrm{\wh w - {\wh w'}}^2. \numberthis \label{eq:w_update_v3}
\end{align*}
Since, the function $\wh F_2$ is $\lambda$-strongly convex, we have that $\nrm{ \grad^2 \wh F_2(\hw) v} \geq \lambda \nrm{v}$ for any vector $v$. Using this fact in \pref{eq:w_update_v3}, we get that 
\begin{align*}
\nrm*{v} \leq \frac{M}{ 2\lambda} \nrm{\wh w - {\wh w'}}^2. 
\end{align*} 
Finally, an application of \pref{lem:w_not_too_far} implies that $\nrm{\wh w - {\wh w'}} \leq \frac{2 m L}{\lambda n}$, using which in the above bound, we get that 
\begin{align*}
\nrm*{v} &\leq \frac{2 M m^2 L^2}{\lambda^3 n^2}. 
\end{align*}
Plugging in the definition of the vector $v$ from \pref{eq:w_update_v7}, we get that 
\begin{align*}
\nrm[\big]{\wh w'  - \wh w - \frac{1}{n - m} \prn{\grad^2 \wh F_2 (\wh w)}^{-1} \sum_{z \in u} \grad f(\wh w; z)} \leq \frac{2 M m^2 L^2}{\lambda^3 n^2}. 
\end{align*}
The desired bound follows by setting $\wh H \ldef{} \frac{1}{n - m} \sum_{z \in S \setminus U} \grad^2 f(\wh w, z) = \grad^2 \wh F_2(\wh w)$.  
\end{proof}

\subsection{Proof of \pref{thm:sc_main_thm}} \label{app:sc_main_thm} 

Before we delve into the proof of \pref{thm:sc_main_thm}, we first provide in \pref{alg:learningalg_rerm_sco}, the pseudocode for the learning algorithm $A_{sc}$. We also recall the following technical result that provides excess risk guarantees for the empirical risk minimizer when the loss is strongly convex and Lipschitz. 
\begin{lemma}[Claim 6.2 in \citet{shalev2009stochasticb}] 
\label{lem:rerm_works_sc} For any $z \in \cZ$, let $f(w, z)$ be a $L$-Lipschitz and $\lambda$-strongly convex function in the variable $w$. Given any distribution $\cD$, let $S = \crl{z_i}_{i=1}^n$ denote a dataset of $n$ samples drawn independently from $\cD$. Let the point $\wh w$ be defined as $\wh w \ldef{} \argmin_{w} \frac{1}{n} \sum_{i=1}^n f(w, z_i).$ Then, 
\begin{align*}
\En \brk*{F(\wh w) - F(w^*)} &\leq \frac{4 L^2}{\lambda n}, 
\end{align*} 
where the function $F(w) \ldef{} \En_{z \sim \cD} \brk*{f(w, z)}$ and $w^* \in \argmin_{w} F(w)$. 
\end{lemma}

\begin{algorithm}[tb]
\caption{Learning algorithm ($A_{sc}$)} 
\begin{algorithmic}[1]
   \Require Dataset $S: \crl*{z_i}_{i=1}^{n} \sim \cD^{n}$, loss function: $f$. 
   \State Compute $$\wh w \leftarrow \argmin\, \wh F_n(w)  \ldef{} \frac{1}{n} \sum_{i=1}^{n} f(w, z_i).$$  
   \State \textbf{Return} $\prn{\wh w, \grad^2 \wh F(\wh w)}$.  
\end{algorithmic} 
\label{alg:learningalg_rerm_sco}  
\end{algorithm}

We are now ready to prove the statements of \pref{thm:sc_main_thm}. We prove each part in a separate lemma below. The following result provides performance guarantee for the output of the learning algorithm $A_{sc}$. 
\begin{lemma}[Learning guarantee for $A_{sc}$] 
\label{lem:learning_guarantee__sc} 
For any distribution $\cD$, the output $\wh w$ of running \pref{alg:learningalg_rerm_sco} on the dataset $S \sim \cD^n$ satisfies 
\begin{align*} 
	\En_{S \sim \cD^{n}} \brk{ F(\wh w)} - F^* \leq \frac{4 L^2}{\lambda n}, 
\end{align*} where $F^*$ denotes $\min_{w \in \cW} F(w)$. 
\end{lemma} 
\begin{proof}[Proof of \pref{lem:learning_guarantee__sc}] We note that the point $\wh w$ is given by the empirical risk minimizer on the dataset $S$, i.e. 
\begin{align*}
\wh w \leftarrow \argmin_{w}\, \frac{1}{n} \sum_{z \in  S} f(w, z). 
\end{align*}
Since the function $f(w, z)$ is $\lambda$-strongly convex and $L$-Lipschitz, the desired performance guarantee for the ERM point follows from \pref{lem:rerm_works_sc}. 
\end{proof}

Next, we provide performance guarantees for the output of the unlearning algorithm $\bar{A}_{sc}$. 

\begin{lemma} 
\label{lem:unlearning_guarantee__sc}
For any dataset $S$, output $\wh w$ of $A_{sc}(S)$ and set $U$ of $m$ delete requests, the point $\wt w$ returned by \pref{alg:unlearningalg_rerm_sco}  satisfies 
\begin{align*}
	\En \brk{F(\wt w) - F^*} = O\prn[\Big]{\frac{ \sqrt{d} M m^2 L^3}{\lambda^3 n^2 \epsilon} \sqrt{\ln\prn[\big]{1/\delta}} + \frac{4mL^2}{\lambda n}}, 
\end{align*} where the expectation above is taken with respect to the dataset $S$ and noise $\nu$. 
\end{lemma} 

\begin{proof}[Proof of \pref{lem:unlearning_guarantee__sc}] We recall that  
\begin{align*}
\wt w = \wh w +  \frac{1}{n - m} \prn{\wh H}^{-1} \sum_{z \in u} \grad f(\wh w, z) +  \nu,  \numberthis \label{eq:unlearning_guarantee__sc}
\end{align*}
where the vector $\nu \in \bbR^d$ is drawn independently from $\cN(0, \sigma^2 \bbI_d)$ with $\sigma$ given by $\sqrt{2 \ln(\frac{1.25}{\delta})} \cdot \frac{2 M m^2 L^2}{\lambda^3 n^2 \epsilon}$. Thus, 
\begin{align*} 
\En \brk*{F(\wt w) - F(w^*)} &= \En \brk*{F(\wt w) - F(\wh w) + F(\wh w) - F(w^*)} \\ 
&= \En \brk*{F(\wt w) - F(\wh w)} + \En \brk*{ F(\wh w) - F(w^*)} \leq \En \brk*{L \nrm*{ \wt w - \wh w}} + \frac{4 L^2}{\lambda n},   \numberthis \label{eq:unlearning_guarantee__sc2}
\end{align*}
 where the inequality in the last line follows from the fact that the function $F = \En \brk*{f(w, z)}$ is $L$-Lipschitz, and by using \pref{lem:learning_guarantee__sc}. Further, from the relation in \pref{eq:unlearning_guarantee__sc}, we have that 
\begin{align*}
\En \brk*{\nrm{\wt w - \wh w}} &= \En \brk{\nrm{  \frac{1}{n - m} \prn{\wh H}^{-1} \sum_{z \in U} \grad f(\wh w, z) + \nu }} \\
&\overleq{\proman{1}} \En \brk{ \sum_{z \in U} \nrm{  \frac{1}{n - m} \prn{\wh H }^{-1} \grad f(\wh w, z)}} + \En \brk*{\nrm{\nu}} \\
&\overleq{\proman{2}} \sum_{z \in U} \frac{1}{(n - m) \lambda} \En \brk{\nrm{ \grad f(\wh w, z) }}  + \sqrt{ \En \brk*{\nrm{\nu}^2}}, 
\end{align*} where the inequality in $\proman{1}$ follows from an application of the triangle inequality, and the inequality $\proman{2}$ holds because the function $F(w)$ is $\lambda$-strongly convex which implies that $\grad^2 F(\wh w) \succcurlyeq \lambda \bbI_d$, and by an application of Jensen's inequality to bound $\En \brk*{\nrm{\nu}}$. Next, using the fact that the vector $\nu \sim \cN(0, \sigma^2 \bbI_d)$, we get that
\begin{align*}
 \En \brk*{\nrm{\wt w - \wh w}} 
&\leq \frac{1}{(n - m) \lambda} \En \brk[\big]{\sum_{z \in U} \nrm{  \grad f(\wh w, z) }}  + \sqrt{d} \sigma \\ 
&\leq \frac{m L}{(n - m) \lambda} + \sqrt{d} \sigma, 
\end{align*} where the last line holds because $f(w, z)$ is $L$-Lipschitz. 
Using the above bound in \pref{eq:unlearning_guarantee__sc2}, we get  
\begin{align*}
\En \brk*{F(\wt w) - F(w^*)} &\leq \frac{mL^2}{(n-m) \lambda} + \sqrt{d} \sigma L + \frac{4L^2}{\lambda n}. 
\end{align*}
Our final guarantee follows by plugging in the value of $\sigma$ and using the fact that $n = \Omega(m)$. 
\end{proof}

Finally, we show that the algorithms $A_{sc}$ and $\bar{A}_{sc}$ are $(\epsilon, \delta)$-unlearning.  

\begin{lemma}[Unlearning guarantee] 
\label{lem:dp_guarantee} For any distribution $\cD$, dataset $S$ and set of delete requests $U \subseteq S$, the algorithms $A_{sc}$ and $\bar{A}_{sc}$ satisfy the following guarantees for any set $W \subseteq \bbR^d$, 
\begin{enumerate}[label=(\alph*)]
\item $\Pr \prn{\bar{A}_{sc}(U, A_{sc}(S), T(S)) \in W} \leq e^\epsilon \Pr \prn{\bar{A}_{sc}(\emptyset, A_{sc}(S \setminus U), T(S \setminus U)) \in W} + \delta$, and 
\item  $\Pr \prn{\bar{A}_{sc}(\emptyset, A_{sc}(S \setminus U), T(S\setminus U)) \in W} \leq e^\epsilon \Pr \prn{\bar{A}_{sc}(U, A_{sc}(S), T(S)) \in W} + \delta$. 
\end{enumerate}  
\end{lemma}
\begin{proof}[Proof of \pref{lem:dp_guarantee}]  The proof follows along the lines of the proof of the differential privacy guarantee for the Gaussian mechanism (see e.g.,  \citet[Appendix A]{DworkR14}). 

Let $\wh w$ denote the output of the learning algorithm $A_{sc}$ when run on dataset $S$, and let $\wt w$ denote the corresponding output of the unlearning algorithm $\bar{A}_{sc}$ when run with delete requests $U$,  the input model $\wh w$ and data statistics $T(S)$, i.e. $\wh w = A_{sc}(S)$ and $\wt w = \bar{A}_{sc}(U, \wh w, T(S))$. Additionally, let $\bar{w}$ be the local variable defined in \pref{line:bar_def} of $\bar{A}_{sc}$ (see \pref{alg:unlearningalg_rerm_sco}) when computing $\wt w$.  

Similarly, let $\wh w'$ denote the output of the learning algorithm $A_{sc}$ when run on dataset $S \setminus U$, and let $\wt w'$ denote the corresponding output of the unlearning algorithm $\bar{A}_{sc}$ when run with delete requests $\emptyset$,  the input model $\wh w'$ and data statistics $T(S \setminus U)$, i.e. $\wh w' = A_{sc}(S \setminus U)$ and $\wt w' = \bar{A}_{sc}(\emptyset, \wh w', T(S \setminus U))$. Additionally, let $\bar{w}'$ be the local variable defined in \pref{line:bar_def} of $\bar{A}_{sc}$ for this case.

Note that in the algorithm $A_{sc}$, the points  $\wh w$ and $\wh w'$ are computed as: 
\begin{align*} 
&\wh w  =  \argmin_{w} \frac{1}{n} \sum_{z \in S} f(w, z) \qquad \text{and} \qquad \wh w' =  \argmin_{w} \frac{1}{n - m} \sum_{z \in S \setminus U} f(w, z). 
\end{align*}
An application of \pref{lem:w_update_v_main} thus gives us the bound    
\begin{align*}
\nrm[\Big]{\wh w'  - \wh w -  \frac{1}{n - m} \prn{\wh H}^{-1} \sum_{z \in U} \grad f(\wh w, z)} \leq  \frac{2 M m^2 L^2}{\lambda^3 n^2}, 
\end{align*} where the matrix $\wh H$ is defined in \pref{eq:H_hat_def}. 
Using the relation in \pref{eq:bar_def} for the points $\bar{w}$ and $\bar{w'}$, and observing that $\wh w' = \bar{w}'$ since $U = \emptyset$ in the calculation of $\wh w'$, we get that 
\begin{align*}
\nrm{\bar{w}' - \bar{w}} \leq \frac{2 M m^2 L^2}{\lambda^3 n^2} \rdef{} \gamma.   \numberthis \label{eq:dp_guarantee2} 
\end{align*} 
Next, note that in the algorithm $\bar{A}_{sc}$, the points $\wt w$ and $\wt w'$ are computed as $\wt w = \bar{w} + \nu$ and $\wt w' = \bar{w}' + \nu$ respectively, where the noise $\nu \sim \cN(0, \sigma^2 \bbI_d)$ with $\sigma = (\gamma / \epsilon) \cdot \sqrt{2 \ln({1.25 / \delta})}$. Thus, following the same proof as \citet[Theorem A.1]{DworkR14}, with the bound \pref{eq:dp_guarantee2}, we get that for any set $W$, 
\begin{align*}
	\Pr \prn*{ \wt w \in W } &\leq e^\epsilon \Pr \prn*{ \wt w' \in W } + \delta,~ \intertext{and } 
	\Pr \prn*{ \wt w' \in W } &\leq e^{\epsilon} \Pr \prn*{ \wt w \in W } + \delta, 
\end{align*} giving us the desired unlearning guarantee. 
\end{proof} 

\begin{proof}[Proof of \pref{thm:sc_main_thm}] The statements of the theorem follow immediately from \pref{lem:learning_guarantee__sc}, \pref{lem:unlearning_guarantee__sc} and \pref{lem:dp_guarantee} respectively. 
\end{proof}

\section{Unlearning algorithms for convex loss function}   \label{app:convex} 
In this section, we provide learning and unlearning algorithms when $f$ is convex (but not necessarily strongly convex). Similar to the strongly convex setting, we assume that 

\begin{assumption}
\label{ass:f_properties_convex} 
For any $z \in \cZ$, the function $f(w, z)$ is convex, $L$-Lipschitz
and $M$-Hessian Lipschitz with respect to $w$. 
\end{assumption} 
In additional to the above, we also assume the following: 
\begin{assumption} \label{ass:w_assumption} There exists a $w^* \in \argmin_{w \in \cW} F(w)$ such that $\nrm{w^*} \leq B$. 
\end{assumption}

Our algorithms for convex losses are based on algorithms for the strongly convex setting. Given the convex function $f(\cdot, z)$, define the function $\wt f(\cdot, z)$ as  
\begin{align*} 
\wt f(w, z) = f(w, z) + \frac{\lambda}{2} \nrm*{w}^2. 
\end{align*} 
The key observation is that at any $w \in \bbR^d$, the function $\wt f(w, z)$ is $\lambda$-strongly convex, $(L + \lambda \nrm{w})$-Lipschitz, $(H + \lambda)$-smooth and $M$-Hessian Lipschitz in $w$ for any $z$. Clearly, the function $\wt f$ satisfies \pref{ass:f_properties} whenever $w$ is such that $\nrm*{w} \leq L / \lambda$ (see \pref{lem:w_bounded_convex}), and thus we can run algorithms $A_{sc}$ and $\bar{A}_{sc}$ respectively on the function $\wt f$.  

\begin{algorithm}
\caption{Learning algorithm ($A_{c}$)}  
\begin{algorithmic}[1] 
   \Require Dataset $S: \crl*{z_i}_{i=1}^{n} \sim \cD^{n}$, loss function: $f$, regularization parameter: $\lambda$. 
   \State Define $$\wt f(w, z) \ldef{ } f(w, z) + \frac{\lambda}{2} \nrm*{w}^2.$$ 
   \State Run the algorithm $A_{sc}$ on the dataset $S$ with loss function $\wt f$. 
   \State \textbf{return} $(\wh w, \wt H) \leftarrow A_{sc}(S; \wt f)$, where $\wt H \ldef{} \frac{1}{n} \sum_{z \in S} \wt f(\wh w, z).$ 
\end{algorithmic} 
\label{alg:learningalg_sco}  
\end{algorithm} 

\begin{algorithm}
\caption{Unlearning algorithm ($\bar{A}_{c}$)} 
\begin{algorithmic}[1] 
   \Require Delete requests: $U = \crl*{z_j}_{j=1}^{m} \subseteq S$, output of $A_{c}$: $\wh w$, additional statistic $T(S): \wt H$, loss function: $f$, regularization parameter: $\lambda$. 
   \State Define $$\wt f(w, z) \ldef{ } f(w, z) + \frac{\lambda}{2} \nrm*{w}^2.$$ 
   \State Run the algorithm $\bar{A}_{sc}$ for the delete request $U$ with input model $\wt w$ and with loss function $\wt f$. 
   \State \textbf{return} $\wt w \leftarrow \bar{A}_{sc}(S, \wh w, T(S); \wt f)$. 
\end{algorithmic} 
\label{alg:unlearningalg_sco}  
\end{algorithm}

Our learning and unlearning algorithms for the convex loss $f$ simply invoke the algorithms $A_{sc}$ and $\bar{A}_{sc}$ on the function $\wt f$ with an appropriate choice of $\lambda$.  We provide the pseudocode in \pref{alg:learningalg_sco} and \pref{alg:unlearningalg_sco} respectively. 

In order to avoid confusion in this section, for any algorithm $A$, we use the notation $A(S; f)$ to denote the fact that $A$ is run on the loss function $f$ (similarly for $\bar{A}$). Whenever clear from context, we will drop the argument $f$ from the notation. Additionally, we also define $\wt F$ to denote the population loss w.r.t. the loss function $\wt f$, i.e. $\wt F(w) \ldef{} \En_{z \sim \cD} \brk{\wt f(w, z)}$. 

\begin{theorem} \label{thm:convex_main_thm} 
Suppose the loss function $f$ satisfy  \pref{ass:f_properties_convex} and \pref{ass:w_assumption}. Let the dataset $S \sim \cD^n$. Then, 
\begin{enumerate}[label=$(\alph*)$] 
\item The point $\wh w$ returned by running $A_{c}$ on $S$ satisfies 
\begin{align*}  
	\En_{S \sim \cD^{n}} \brk{ F(\wh w) - \min_{w \in \cW} F(w)} \leq\frac{\lambda B^2}{2} +  \frac{16 L^2}{\lambda n}.  \numberthis \label{eq:bound1_convex}  
\end{align*} 

\item For any set $U \subseteq S$ of $m$ delete requests,  the point $\wt w$ returned by $\bar{A}_{c}$ satisfies 
\begin{align*}
	\En_{S, \nu} \brk{F(\wt w) - \min_{w \in \cW} F(w)} = O\Big( \frac{\lambda B^2}{2} +  \frac{\sqrt{d} M m^2 L^3}{\lambda^3 n^2 \epsilon} \sqrt{\ln\prn{1/\delta}} + \frac{m L^2}{\lambda n} \Big). 
 \numberthis \label{eq:bound2_convex}  
\end{align*} 
 
\item The learning algorithm $A_{c}$ and the unlearning algorithm $\bar{A}_{c}$ are $(\epsilon, \delta)$-unlearning. 
\end{enumerate}
\end{theorem} ~

\begin{corollary} Suppose we did not have any unlearning requests, and only cared about the performance of the output point $\wh w$ for the learning algorithm $A_c$. Then, setting $\lambda = L / B \sqrt{n}$, the performance guarantee for the point $\wh w$ given in \pref{thm:convex_main_thm} implies  
\begin{align*}
\En \brk{F(\wh w) - F^*} &\leq \frac{BL}{\sqrt{n}}.  
\end{align*} The above rate is tight for learning with Lipschitz convex losses (see  \citet[Theorem 6.1]{bubeck2014convex}).  
\end{corollary}

\begin{corollary} 
\label{corr:corollary2}
Suppose that we have $m$ delete requests and thus care about the performance guarantee of both the point $\wh h$, output of the learning algorithm $A_{c}$, and the point $\wt w$, output of the unlearning algorithm $\bar{A}_{c}$. In this case, we set the regularization parameter $\lambda$ as: 
\begin{align}
\lambda = \max \crl[\Big]{\frac{L}{B} \sqrt{\frac{m}{n}}, \prn[\Big]{{\frac{ \sqrt{d} M m^2 L^3}{B^2 n^2 \epsilon} \sqrt{\ln\prn[\big]{1/\delta}}}}^{1/4}}. \numberthis \label{eq:setting_lambda} 
\end{align} 
Plugging the above values of $\lambda$ in \pref{thm:convex_main_thm}, we get that 
\begin{align*}
\small \En \brk*{F(\wh w) - F^*} = O\prn[\Big]{c_1\sqrt{\frac{m}{n}} + c_2 \prn[\Big]{\frac{d \log(1/\delta)}{\epsilon^2}}^{1/8} \sqrt{\frac{m}{n}}} \\
\intertext{and}
\small \En \brk*{F(\wt w) - F^*} = O\prn[\Big]{c_1 \sqrt{\frac{m}{n}} + c_2 \prn[\Big]{\frac{d \log(1/\delta)}{\epsilon^2}}^{1/8} \sqrt{\frac{m}{n}}},  
\end{align*} 
where the constant $c_1 \propto BL$ and $c_2 \propto \prn[\Big]{ \frac{M L^3}{B^2} }^{1/4}$. 
\end{corollary}

\subsection{Proof of  \pref{thm:convex_main_thm}}
We are now ready to prove the statements of \pref{thm:convex_main_thm}. We prove each part in a separate lemma below. The following  provides performance guarantee for the output of the learning algorithm $A_{c}$.

\begin{lemma}[performance guarantee for $A_{c}$]
\label{lem:convex_learning}
For any $\lambda > 0$ and $S \sim \cD^n$, the point $\wh w$ returned by \pref{alg:learningalg_sco} satisfies
\begin{align*}
\En \brk{F(\wh w) - F^*} &\leq \frac{\lambda B^2}{2} +  \frac{16 L^2}{\lambda n}.  
\end{align*} 
\end{lemma}
\begin{proof} First, note that an application of \pref{lem:w_bounded_convex} implies that for any dataset $S$, the empirical minimizer $\wh w$ (returned by $A_c$) satisfies: $\nrm*{\wh w} \leq L / \lambda$. Thus, our domain of interest is $\cW \ldef{} \crl{w \mid \nrm{w} \leq L / \lambda}$. Over the set $\cW$, the function $\wt f$ is $2L$-Lipschitz, and thus, an application of \pref{lem:learning_guarantee__sc}  implies that the returned point $\wh w$ satisfies 
\begin{align*}
\En \brk{\wt F(\wh w)} &\leq \wt F(\wt w^*) + \frac{16 L^2}{\lambda n}, 
\end{align*} where $\wt w^*$ denotes the minimizer of $\wt F(w)$. We can further upper bound the right hand side above as 
\begin{align*}
\En \brk{\wt F(\wh w)} &\leq \wt F(w^*) + \frac{16 L^2}{\lambda n}, 
\end{align*}
where $w^*$ denotes a minimizer of the population loss $F(w)$ that satisfies \pref{ass:w_assumption}. Plugging in the form of the function $\wt F(w)$  in the above, we get  
\begin{align*}
\En \brk*{F(\wh w)} &\leq F(w^*) + \frac{\lambda}{2} \nrm*{w^*}^2 +  \frac{16 L^2}{\lambda n} \\
			   &\leq  F(w^*) + \frac{\lambda B^2}{2} +  \frac{16 L^2}{\lambda n}, 
\end{align*} where the last line holds due to \pref{ass:w_assumption}. 
\end{proof}

\begin{lemma}[performance guarantee for $\bar{A}_{c}$] \label{lem:convex_unlearning} 
For any $\lambda > 0$, dataset $S \sim \cD^n$, output $\wh w$ of $A_{c}(S)$ and set $U$ of $m$ delete requests, the point $\wt w$ returned by \pref{alg:unlearningalg_sco} satisfies 
\begin{align*} 
\En \brk{ F(\wh w)- F^*} &= O\Big( \frac{\lambda B^2}{2} +  \frac{\sqrt{d} M m^2 L^3}{\lambda^3 n^2 \epsilon} \sqrt{\ln\prn{1/\delta}} + \frac{m L^2}{\lambda n} \Big). 
\end{align*}
\end{lemma} 
\begin{proof} Let $w^* \in \argmin_{w} F(w)$. We note that  
\begin{align*}
\En \brk*{F(\wt w) - F(w^*)} &= \En \brk*{F(\wt w) - F(\wh w) + F(\wh w) - F(w^*)} \\ 
&= \En \brk*{F(\wt w) - F(\wh w)} + \En \brk*{ F(\wh w) - F(w^*)} \\   
&\leq \En \brk*{L \nrm*{ \wt w - \wh w}} + \En \brk*{ F(\wh w) - F(w^*)},   \numberthis \label{eq:unlearning_convex_sc2} 
\end{align*} where the inequality in the last line holds because the loss function $f(w, z)$, and thus the function $F(w)$, is $L$-Lipschitz. We next note that $\bar{A}_{c}$ computes the point $\wt w$ by running the algorithm $\bar{A}_{sc}$ with inputs $U$, $\wh w$ on the loss function $\wh f$. Thus, we have that 
\begin{align*}
 \hspace{0.3in} & \hspace{-0.3in} \En \brk*{\nrm{\wt w - \wh w}} = \En \brk[\Big]{\nrm[\Big]{  \frac{1}{n - m} \prn{\wh H}^{-1} \sum_{z \in U} \grad \wt f(\wh w, z) + \nu }}, 
\end{align*} where $\wh H \ldef{} \frac{1}{n -m} \sum_{z \in S \setminus U} \grad^2 \wt f(w, z)$. Since, the function $\wt f$ is $\lambda$-strongly convex, we note that $\wh H  \succcurlyeq \lambda \bbI_d$. Using this fact with the above relation implies 
\begin{align*}
 \En \brk*{\nrm{\wt w - \wh w}} &\leq \frac{1}{\lambda(n -m)} \nrm{ \sum_{z \in U}  \grad \wt f(\wh w, z)} + \En \nrm*{\nu} \\ 
 &\leq \frac{1}{\lambda(n -m)} \sum_{z \in U} \nrm{ \grad \wt f(\wh w, z)} + \En \nrm*{\nu} \leq \frac{2mL}{\lambda(n - m)} +  \sigma,  \numberthis \label{eq:unlearning_convex_sc3}
\end{align*} where the last line follows from the fact that $\nrm*{\wh w} \leq \frac{L}{\lambda}$, and thus $$\nrm{\grad \wt f(\wh w, z)} \leq \nrm*{\grad f(\wh w, z)} + \lambda \nrm{\wh w} \leq 2L.$$ 

Finally, using \pref{eq:unlearning_convex_sc3} and  \pref{lem:convex_learning} in  \pref{eq:unlearning_convex_sc2}, and by plugging in the value of $\sigma$ implies the desired performance guarantee. 
\end{proof}

\noindent 
Finally, the algorithms $A_c$ and $\bar{A}_c$ are $(\epsilon, \delta)$-forgetting, as a consequence of \pref{lem:dp_guarantee}. 

\begin{lemma}[Forgetting guarantee] 
\label{lem:dp_guarantee_convex} For any distribution $\cD$, dataset $S$, set of delete requests $U \subseteq S$, the algorithms $A_{c}$ and $\bar{A}_{c}$ satisfy the following guarantees for any set $W \subseteq \bbR^d$, 
\begin{enumerate}[label=(\alph*)]
\item $\Pr \prn{\bar{A}_{c}(U, A_{c}(S), T(S)) \in W} \leq e^\epsilon \Pr \prn{\bar{A}_{c}(\emptyset, A_{c}(S \setminus U), T(S \setminus U)) \in W} + \delta$, and 
\item  $\Pr \prn{\bar{A}_{c}(\emptyset, A_{c}(S \setminus U), T(S\setminus U)) \in W} \leq e^\epsilon \Pr \prn{\bar{A}_{c}(U, A_{c}(S), T(S)) \in W} + \delta$. 
\end{enumerate}  
\end{lemma}
\begin{proof}[Proof of \pref{thm:convex_main_thm}] The desired statements follow immediately from \pref{lem:convex_learning}, \pref{lem:convex_unlearning} and \pref{lem:dp_guarantee_convex} respectively. 
\end{proof}

\subsection{Supporting technical results}
The following lemma gives a bound on the regularized empirical risk minimizer point for the loss function $f(w, z)$ for any dataset $S$. This gives us a bound on the domain of interest, and thus allows us to bound the Lipschitz constant for loss function over this domain. 

\begin{lemma} 
\label{lem:w_bounded_convex}
Let $f(w, z)$ be a $L$-Lipschitz function in the variable $w$, and let $\wt f$ be defined as 
\begin{align*}
	\wt f(w, z) = f(w, z) + \frac{\lambda}{2} \nrm*{w}^2. 
\end{align*} 
Given a dataset $S = \crl*{z_i}_{i=1}^n$, define $\wh G(w) = \frac{1}{n} \sum_{i=1}^n \wt f(w, z_i)$, and let $\wh w$ denote the empirical risk minimizer of the loss function $\wt f$ on dataset $S$, i.e. $\wh w \in \argmin_{w} \wh G(w)$. Then, the point $\wh w$ satisfies $\nrm{\wh w} \leq \frac{L}{\lambda} 
$.
\end{lemma}
\begin{proof} Since, $\wh w \in  \argmin_{w} \wh G(w)$, we have 
\begin{align*}
\grad \wh G(\wh w) =  \frac{1}{n} \sum_{z \in S} \grad \wt f(\wh w, z) = 0. 
\end{align*}
Plugging in the definition of the function $\wt f$ in the above implies that \begin{align*}
\frac{1}{n} \sum_{z \in S} \grad f(\wh w, z) + \lambda \wh w = 0. 
\end{align*}
Rearranging the terms, we get 
\begin{align*}
\nrm{\lambda \wh w} &= \nrm{\frac{1}{n} \sum_{z \in S} \grad f(\wh w, z) } \overleq{\proman{1}} \frac{1}{n} \sum_{z \in S} \nrm{\grad f(\wh w, z) } \overleq{\proman{2}} L, 
\end{align*} where the inequality $\proman{1}$ follows from an application of the Triangle inequality, and the inequality in $\proman{2}$ holds because the function $f$ is $L$-Lipschitz in the variable $w$, and thus $\nrm*{\grad f(w, z)} \leq L$ for all $w \in \cW$ and $z \in \cZ$.  
\end{proof}

\end{document}